\documentclass{article}

\usepackage{microtype}
\usepackage{graphicx}
\usepackage{subfigure}
\usepackage{booktabs} 

\usepackage{hyperref}
\usepackage{hyperref,url,booktabs,amsfonts,nicefrac,microtype, xcolor, graphicx,amsmath,floatrow, subfig, wrapfig, enumerate, makecell, multirow, stackengine,wasysym,scalerel, caption, bbm}

\usepackage{subcaption}
\usepackage{caption}

\newlength\savewidth

\newcolumntype{x}[1]{>{\centering\arraybackslash}p{#1pt}}
\newcolumntype{y}[1]{>{\raggedright\arraybackslash}p{#1pt}}
\newcolumntype{z}[1]{>{\raggedleft\arraybackslash}p{#1pt}}
\usepackage{float}

\newcommand{\app}{\raise.17ex\hbox{$\scriptstyle\sim$}}

\newfloatcommand{capbtabbox}{table}[][\FBwidth]

\usepackage[accepted]{icml2025}

\usepackage{amsmath}
\usepackage{amssymb}
\usepackage{mathtools}
\usepackage{amsthm}
\usepackage{placeins} 
\usepackage[capitalize,noabbrev]{cleveref}

\theoremstyle{plain}
\newtheorem{theorem}{Theorem}[section]

\theoremstyle{definition}
\newtheorem{definition}[theorem]{Definition}

\theoremstyle{remark}

\expandafter\def\expandafter\normalsize\expandafter{%
    \normalsize%
    \setlength\abovedisplayskip{0pt}%
    \setlength\belowdisplayskip{8pt}%
    \setlength\abovedisplayshortskip{-8pt}%
    \setlength\belowdisplayshortskip{2pt}%
}

\usepackage[textsize=tiny]{todonotes}

\newcommand{\TODO}[1]{{}} 

\begin{document}

\twocolumn[

\icmltitle{Do we need equivariant models for molecule generation?}

\begin{icmlauthorlist}
\icmlauthor{Ewa M. Nowara}{pd-sf}
\icmlauthor{Joshua Rackers}{achira}
\icmlauthor{Patricia Suriana}{pd-sf} \\
\icmlauthor{Pan Kessel}{pd-bas}
\icmlauthor{Max Shen}{pd-sf}
\icmlauthor{Andrew Martin Watkins}{pd-sf}
\icmlauthor{Michael Maser}{pd-sf}

\end{icmlauthorlist}
\vskip 0.3in

\icmlaffiliation{pd-sf}{Prescient Design, Genentech, South San Francisco, USA}
\icmlaffiliation{pd-bas}{Prescient Design, Roche, Basel, Switzerland}
\icmlaffiliation{achira}{Achira}

\icmlcorrespondingauthor{Ewa M. Nowara}{nowara.ewa@gene.com}

\icmlkeywords{Machine Learning, Equivariant Models, Drug Discovery, Generative Models, 3D Generation, Voxel Structures, Molecules}
]



\printAffiliationsAndNotice{}  

\begin{abstract}	
Deep generative models are increasingly used for molecular discovery, with most recent approaches relying on equivariant graph neural networks (GNNs) under the assumption that explicit equivariance is essential for generating high-quality 3D molecules. However, these models are complex, difficult to train, and scale poorly. 

We investigate whether non-equivariant convolutional neural networks (CNNs) trained with rotation augmentations can learn equivariance and match the performance of equivariant models. We derive a loss decomposition that separates prediction error from equivariance error, and evaluate how model size, dataset size, and training duration affect performance across denoising, molecule generation, and property prediction. To our knowledge, this is the first study to analyze learned equivariance in generative tasks.
\end{abstract}

\section{Introduction}
\label{Introduction}

Generative models have transformed fields like computer vision and natural language processnig, and are increasingly applied in drug discovery to design novel molecules. Many recent approaches rely on equivariant deep learning, particularly equivariant graph neural networks (GNNs)~\cite{hoogeboom2022equivariant, gebauer2019symmetry, vignac2023midi, EEG-SDE, huang2024learning, xu2023geometric}, based on the assumption that equivariance to 3D rotations and translations is essential for high-quality molecular structures. However, enforcing equivariance through architectural design introduces practical limitations: these models are harder to scale, slower to train, and more difficult to implement.

An alternative is to use data augmentation, applying random rotations and translations during training, which encourages learned equivariance without architectural constraints. This allows the use of more flexible architectures such as 3D convolutional neural networks (CNNs). Recent CNN-based models~\cite{pinheiro20233d, pinheiro2024structure, nowara2024nebula} have outperformed equivariant GNNs, despite lacking explicit equivariance.

We investigate whether CNNs trained with data augmentation learn an equivariant behavior, and how model size, dataset size, and training time influence it. We study this across three tasks: molecule reconstruction from noise (denoising), molecule generation, and property prediction using the public GEOM-Drugs dataset~\cite{geomdrugs_axelrod2022geom}.

While equivariance has been studied extensively in supervised tasks (e.g., classification, segmentation~\cite{gerken2024emergent, quiroga2020revisiting, gerken2022equivariance}), its role in generative modeling remains underexplored. To our knowledge, this is the first systematic evaluation of learned equivariance in molecular generative models. We focus on \textit{seeded} generation (starting from a given lead compound), where equivariance is especially critical. If outputs vary under input rotations, the model becomes unreliable and computationally inefficient.

We find that CNNs trained with rotation augmentation easily learn equivariance during denoising—even with small models, limited data (as little as 10\%), and few training epochs. Property prediction shows similar robustness. However, in generative tasks, only larger models trained on more data produce consistent outputs across rotated seeds. Smaller models diverge, losing generative equivariance.

Interestingly, latent embeddings of rotated molecules differ significantly, even when reconstructions and predictions are nearly identical, suggesting that CNNs learn redundant parallel representations instead of recognizing rotated inputs as equivalent. This inefficiency may explain their higher parameter demands compared to equivariant GNNs. We hypothesize that auxiliary training objectives used to align latent embeddings across rotations could improve robustness without requiring large models or datasets.

\section{Related Work}
\subsection{Equivariance in Machine Learning} 
Several works compare data augmentation and equivariant architectures. See~\citet{bronstein2021geometric} for an overview of geometric deep learning and~\citet{fei2024rotation} for a survey of equivariance in 3D machine learning. \citet{kvinge2022ways} introduced metrics for evaluating invariance and equivariance in both approaches. \citet{gerken2024emergent} showed that equivariance can emerge from data augmentation in predictive tasks, and~\citet{quiroga2020revisiting} found no advantage in using equivariant CNNs for image classification. However, \citet{gerken2022equivariance} showed that equivariant models significantly outperform augmented CNNs on spherical image segmentation, which they argue is a more challenging setting. They also found that larger models and datasets help, but are not sufficient to match equivariant architectures. Our work is the first to study how learned equivariance affects the distribution of generated samples in generative models.

\subsection{Molecule Generation} 
3D molecular generation improves over 2D representations by capturing spatial structure. Existing models typically use either point cloud representations with explicitly equivariant GNNs, or voxel-based CNNs with data augmentation.

GSchNet~\cite{gebauer2019symmetry} introduced point-set autoregressive generation. EDM~\cite{hoogeboom2022equivariant} used SE(3)-equivariant diffusion to denoise point clouds from Gaussian noise. MiDi~\cite{vignac2023midi} jointly generated molecular graphs and 3D conformations.

Voxel-based models use CNNs for generation. Early works~\cite{skalic2019shape, ragoza2020learning} applied VAEs~\cite{kingma2013auto} to voxelized inputs. VoxMol~\cite{pinheiro20233d} used walk-jump sampling (WJS)~\cite{saremi2019neural}, a diffusion-inspired process, for efficient generation. Nebula~\cite{nowara2024nebula} trained a latent generative model on voxel embeddings to improve generalization. Voxel models have also been extended for structure-based generation conditioned on a protein pocket~\cite{pinheiro2024structure}.

\section{Background}

\subsection{Equivariance}

We begin by defining equivariance and approximate equivariance that arises in models trained with data augmentation.

\begin{definition}[Equivariant Functions] A function $f$ is equivariant to a transformation $R$ if applying $R$ to the input $x$ corresponds to applying the transformation $R$ to the output, such that \begin{equation} f(R(x)) = R(f(x)) \end{equation} \end{definition}

In this work, we focus on SE(3) equivariance, encompassing both rotations and translations. SE(3) equivariance excludes reflections, which can alter molecular properties, so we typically do not want models to be equivariant to them. Since convolutional neural networks (CNNs) are translationally equivariant by design, we primarily study rotational equivariance.

\begin{theorem}[Loss decomposition for quantifying learned equivariance]
\label{theorem:loss_decomp}
For a function $f: \mathbb{R}^D \rightarrow \mathbb{R}^D$, a loss function $l: \mathbb{R}^D \times \mathbb{R}^D \rightarrow \mathbb{R}$, a dataset $X$ with labels $Y$, and a class of transformations $R$,
\end{theorem}

\begin{equation}
\begin{aligned}
    \mathbb{E}_{x,R} [ l(f(R(x)), R(y)) ]
    &\approx
    \underbrace{
        \mathbb{E}_x \left[ l\left( \mathbb{E}_R[ R^{-1} f(R(x)) ], y \right) \right]
    }_{\text{prediction error}} \\
    &\quad +
    \underbrace{
        \frac{1}{2} \mathbb{E}_x \left[
            \text{tr}\left(
                H_l(\mu, y)
                \, \text{Var}_R( R^{-1} f(R(x)) )
            \right)
        \right]
    }_{\text{equivariance error}}
\end{aligned}
\end{equation}

where $H_l(\mu, y)$ is the Hessian of the loss function $l$ given terms $\mu = \mathbb{E}_R[ R^{-1} f(R(x)) ]$ and $y$, and $\text{Var}$ is a $D \times D$ covariance matrix.

Moreover, if $l$ is MSE loss, then the relation holds exactly, and simplifies to:

\begin{equation}
\begin{aligned}
\mathbb{E}_{x,R} \left[ l(f(R(x)), R(y)) \right]
&= \underbrace{
    \mathbb{E}_x \left[ l\left( \mathbb{E}_R\left[ R^{-1} f(R(x)) \right], y \right) \right]
}_{\text{prediction error}} \\
&\quad + \underbrace{
    \frac{1}{D} \mathbb{E}_x \left[
        \sum_{i=1}^D
            \text{Var}_R\left( \left[ R^{-1} f(R(x)) \right]_i \right)
    \right]
}_{\text{equivariance error}}
\end{aligned}
\end{equation}

\begin{proof}
This follows by a second-order Taylor expansion around $\mu$.
\end{proof}

\begin{definition}[Approximate Equivariance] A model trained with rotation augmentation can learn approximate equivariance, even if its architecture is not explicitly equivariant. \end{definition}

For a denoising model, motivated by theorem \ref{theorem:loss_decomp}, we quantify rotational equivariance by comparing the reconstruction of a rotated input molecule $\hat{x}_n$ to the rotated reconstruction of the unrotated input $R(\hat{x}_0)$. An equivariant model should satisfy:

\begin{equation} || R(\hat{x}_0) - \hat{x}_n ||_2^2 < \epsilon \label{eq:reconstr} \end{equation}

where $\epsilon$ is a small threshold corresponding to the reconstruction error floor, and squared L2-norm is related to the variance term in theorem \ref{theorem:loss_decomp}.

\subsection{Equivariance of the Denoising Process}

Let $x$ denote clean data (e.g., a molecule), and $y$ the corresponding noisy input after adding isotropic Gaussian noise. The noisy data distribution is:

\begin{equation} p(y) = \int p(x)p(y|x)dx \end{equation}

Assuming the data distribution and the noise process are both rotation-equivariant: \begin{equation} \begin{aligned} p(Rx) &= p(x) \\ p(Ry|Rx) &= p(y|x) \end{aligned} \end{equation}

Then, we can show that the smoothed noisy distribution $p(y)$ is also equivariant if smoothing and data density is also equivariant, by substituting $x = R \tilde{x}$.

\begin{equation}
\begin{aligned}
p(Ry) &= \int p(x)\, p(Ry\,|\,x)\, dx \\
      &= \int p(R\tilde{x})\, p(Ry\,|\,R\tilde{x})\, d\tilde{x} \quad \text{(where } x = R\tilde{x}) \\
      &\text{(since } \det R = 1 \text{, we have } dx = d(R\tilde{x}) = d\tilde{x}) \\
      &= \int p(\tilde{x})\, p(y\,|\,\tilde{x})\, d\tilde{x} \\
      &= p(y)
\end{aligned}
\end{equation}

Let $D$ be the denoising function. The denoised distribution is: \begin{equation}
\hat{p}(x) = \int \delta(x - D(y))\, p(y)\, dy
\end{equation}

If the denoiser is equivariant, i.e., $D(Ry) = R D(y)$, and $p(Ry) = p(y)$ as above, then, by substituting $y = R \tilde{y}$.


\begin{equation}
\begin{aligned}
\hat{p}(x) &= \int \delta(Rx - D(y))\, p(y)\, dy \\
           &= \int \delta(Rx - D(R\tilde{y}))\, p(R\tilde{y})\, d\tilde{y} \\
           &= \int \delta(R(x - D(\tilde{y})))\, p(\tilde{y})\, d\tilde{y} \\
           &= \hat{p}(x)
\end{aligned}
\end{equation}

Using the fact that $\delta(R(x - D(y))) = \delta(x - D(y))$ for any invertible matrix $R$ with $\det R = 1$ which holds for all $R \in \mathrm{SO}(3)$ the delta function is invariant under such transformations. Thus, the denoised distribution is rotation-equivariant if both the noise and the denoiser are equivariant.

\subsection{Molecule Generation}

\textbf{Representing Molecules as 3D Voxels.} Following~\cite{pinheiro20233d}, we represent each molecule as 3D voxels by drawing a continuous Gaussian density around each atom's atomic coordinates on a voxel grid and represent each atom type as a separate channel, resulting in a 4D tensor of $[c \times l \times l \times l]$ for each molecule, where $c$ is the number of atom types and $l$ is the length of the voxel grid edge. The voxel values are normalized to a 0 to 1 range. See Fig~\ref{fig:reconstr_setup} for an example of a voxel representation of a molecule.

\textbf{Generating Molecules with Walk Jump Sampling.} We generate new molecules with a sampling process called \textit{neural empirical} Bayes~\cite{saremi2019neural}. It is a two-step score-based sampling method, called \textit{walk-jump sampling} (WJS). In the first ``walk" step, we sample new noisy molecular voxel grid with multiple k walk steps of a Langevin Markov chain Monte Carlo (MCMC) sampling~\cite{cheng2018underdamped} along a randomly initialized manifold (eq.~\ref{eq:8}). In the second ``jump" step, we denoise the newly sampled noisy molecule grid with a forward pass of a pre-trained denoising autoencoder (DAE) model at an arbitrary step k (eq.~\ref{eq:9}). The DAE is trained to denoise voxelized molecules with random isotropic Gaussian noise added to their voxel grid with a mean squared error (MSE) loss between ground truth and reconstructed voxels. WJS is very similar to diffusion models but it is much faster as it only requires a single noising and denoising step~\cite{pinheiro20233d, nowara2024nebula, pinheiro2024structure}.

\begin{equation}
\begin{split}
\begin{aligned}
d\upsilon_t &= - \gamma \upsilon_t dt - ug_{\theta} (y_{t}) dt + (\sqrt{2\gamma u}) dB_t, \\
d y_{t} &= \upsilon_t dt,
\label{eq:8}
\end{aligned}
\end{split}
\end{equation}

where $B_t$ is Brownian motion, $\gamma$ is friction, $u$ is inverse mass, and $\gamma$, $u$ are hyperparameters, and $g_{\theta}$ is the learned denoising score function.

\begin{equation}
\hat{x} = g_{\theta}(y)
\label{eq:9}
\end{equation}

To obtain a molecular graph from the generated voxels, we localize the atomic coordinates as the center of each voxel using an optimization-based peak finding method~\cite{pinheiro20233d}. 

\section{Experiments and Results}

\subsection{Reconstruction}

\begin{figure}[htbp!]

\centering
\centerline{\includegraphics[width=\columnwidth]{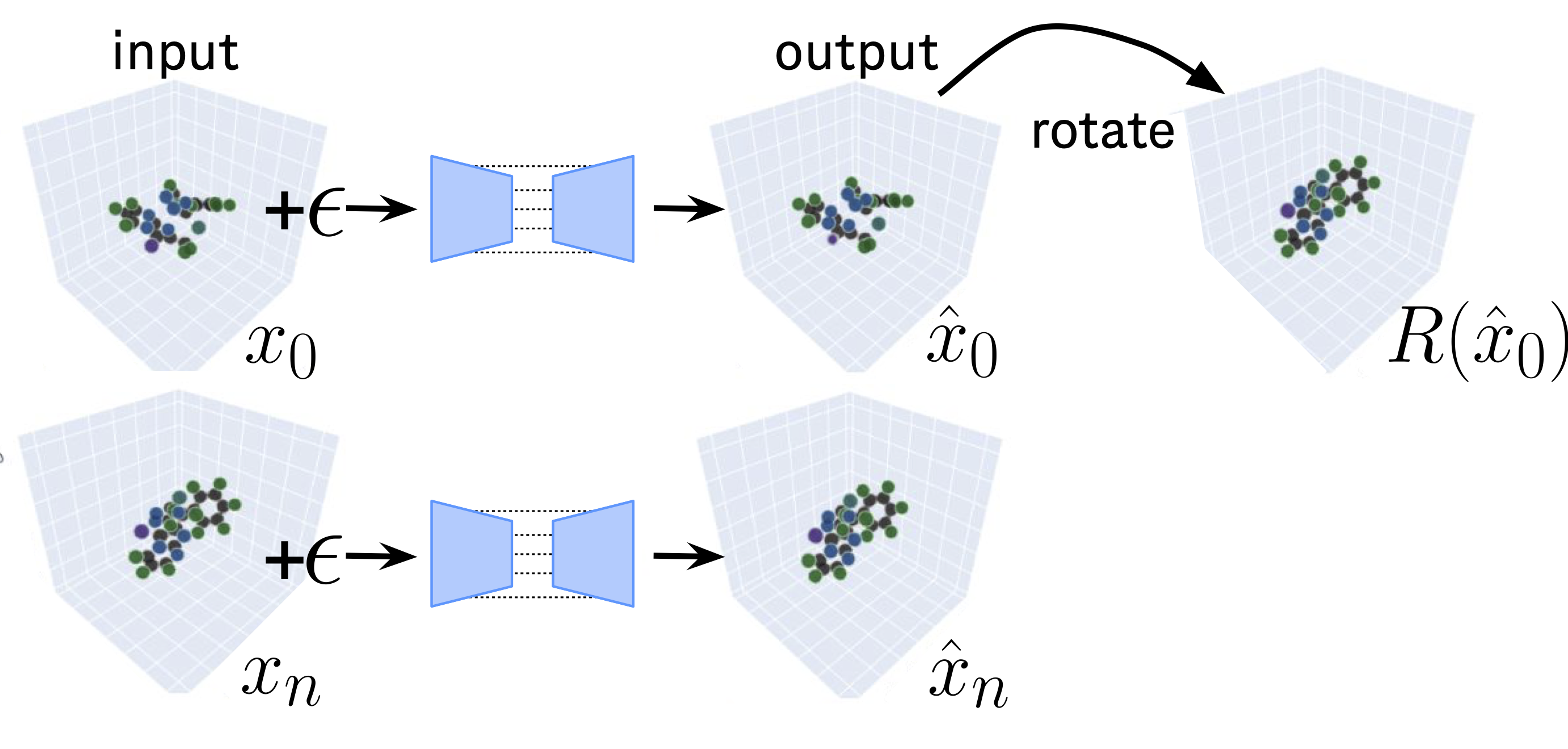}}
\caption{Experimental set up of reconstruction equivariance to rotations of voxelized molecules. We compute the reconstruction error between reconstructed molecules from a rotated input $\hat{x}_n$ to the rotated reconstruction of an unrotated input $R(\hat{x}_0)$.}
\label{fig:reconstr_setup}
\vskip -0.2in
\end{figure}

\begin{figure}[htbp!]

\centering
\centerline{\includegraphics[width=\columnwidth]{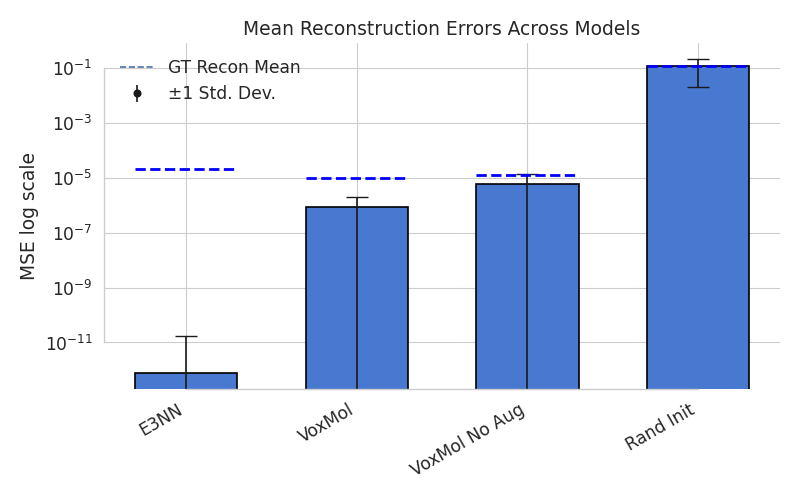}}
\caption{Rotational equivariance with different models compared to reconstruction upper bound computed between reconstructed and ground truth molecule (dotted blue lines).}
\label{fig:reconstr_models}
\vskip -0.2in
\end{figure}

Generative models are trained to denoise the input from Gaussian noise~\cite{pinheiro20233d, nowara2024nebula, rombach2022high}. If a model is unable to properly reconstruct (denoise) the input, the generation quality will be poor. Therefore, first we study the quality of the molecule reconstruction and how it's affected by molecule rotations.

To evaluate whether generative models learn rotational equivariance, we measure reconstruction equivariance to rotations of the input molecule. By definition of equivariance in eq.~\ref{eq:reconstr}, we compare the reconstructed molecule from a rotated input $\hat{x}_n$ to the rotated reconstruction of an unrotated input $R(\hat{x}_0)$. A model is equivariant if these outputs are identical or differ only by a small, bounded error. Figure~\ref{fig:reconstr_setup} shows the experimental set up for reconstruction equivariance. 

We compare VoxMol~\cite{pinheiro20233d} trained with rotation and translation augmentations and its equivariant version (E3NN). Due to prohibitively long training times of the E3NN, we were only able to train a smaller model with 0.5M parameters compared to 111M in VoxMol. The training cost of E3NN per epoch was 90 minutes compared to only 2 hours for the full size VoxMol which has 200 times more parameters.

Note that we use the same amount of training data for VoxMol trained with and without data augmentation. Rather than increasing the dataset size by presenting each molecule in multiple rotated forms, we apply a random rotation to each molecule once per training iteration. This preserves the effective training set size while introducing rotational variability. Detailed results with reconstruction errors for each rotation axis and angle are reported for each experiment in the Appendix~\ref{sec:appendix_rec}. See Appendix~\ref{sec:appendix_arch} for architecture, training details, and dataset~\ref{sec:appendix_dataset}.

Figure~\ref{fig:reconstr_models} shows that both equivariant E3NN and VoxMol trained with rotation augmentations achieve low equivariant reconstruction error. Notably, these errors are consistently lower than the ground truth reconstruction error (between the input and the output in 3D) which can be regarded as an upper bound of reconstruction capability of a model, indicating strong approximate equivariance even for models without explicitly equivariant architectures. While E3NN achieves a lower reconstruction error and lower equivariance error than VoxMol, it does not translate to better molecule generation quality, as will be demonstrated in the next section.

\textbf{Training with No Augmentations.} Interestingly, VoxMol trained without rotation augmentation also demonstrates a lower equivariant error than ground truth reconstruction error, suggesting that the 3D CNN architecture and the denoising task inherently capture some rotational structure from the sparse voxel inputs. 

\textbf{Random Initialization.} To verify whether equivariance arises automatically from the architecture or from the sparsity of the voxel representation, we evaluate the reconstructions with a randomly initialized VoxMol with no training. However, we find that the untrained model does not exhibit reconstruction equivariance although it can be learned very quickly during training. This confirms that equivariance is not an inherent property of the architecture or the voxel inputs, but rather a learned behavior that emerges during training with rotation augmented data.

\begin{figure}[htbp!]

\centering
\centerline{\includegraphics[width=\columnwidth]{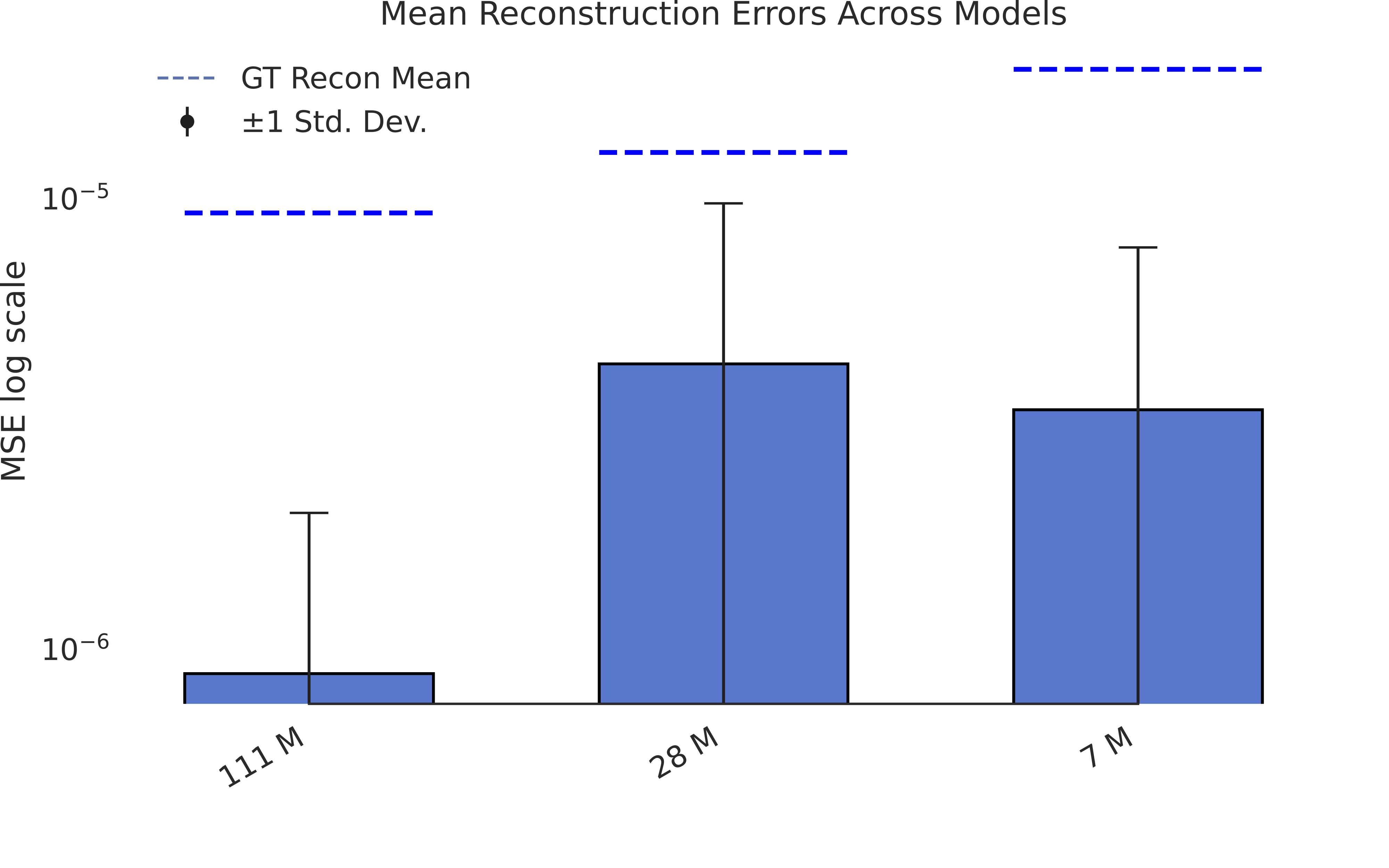}}
\caption{Effect of \textbf{model size} on reconstruction equivariance. Even much smaller models are able to learn equivariance during reconstruction.}
\label{fig:reconstr_model_size}
\vskip -0.2in
\end{figure}

\textbf{Effect of Model Size.} We also evaluate VoxMol models with significantly fewer parameters 28 million (M) and 7 M, compared to the full 111 M model in Figure~\ref{fig:reconstr_model_size}. Despite reduced reconstruction accuracy, both smaller models exhibit similarly lower equivariant reconstruction error compared to the upper bound. This suggests that model size has little impact on a model’s ability to learn equivariance from rotation augmentation, even though it does affect the overall quality of reconstruction and consequently will affect the quality of the generations. 

\begin{figure}[htbp!]

\centering
\centerline{\includegraphics[width=\columnwidth]{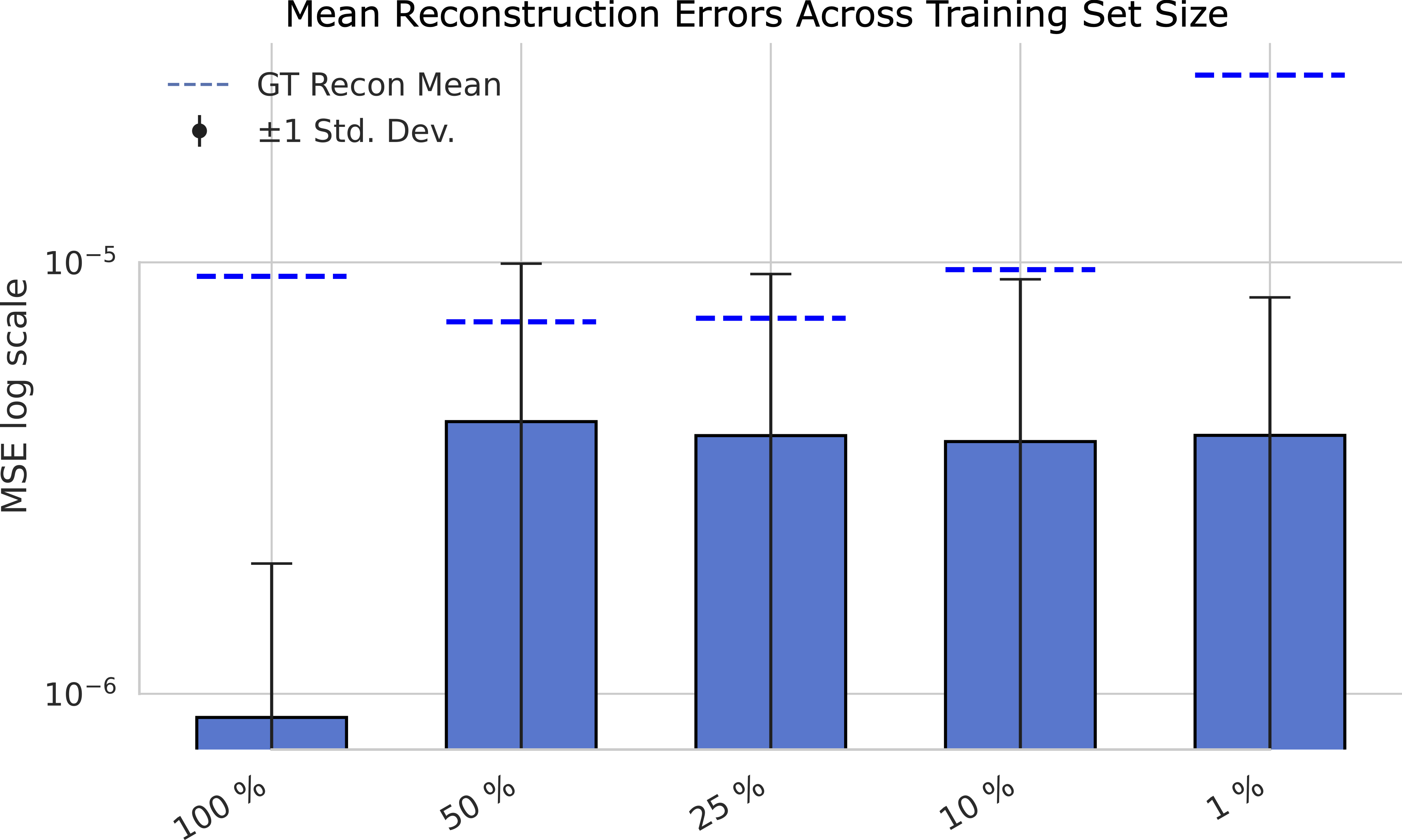}}
\caption{Effect of \textbf{training set size}. Equivariance is already learned with limited data although the overall reconstruction errors increase with decreasing amount of training data.}
\label{fig:reconstr_train_size}
\vskip -0.2in
\end{figure}

\textbf{Effect of Training Set Size.} Prior work has suggested that large datasets and long training times are needed for the learned equivariance to emerge~\cite{gerken2022equivariance}. Contrary to that belief, we find that VoxMol learns strong reconstruction equivariance even when trained on as little as 1\% of the dataset. Figure~\ref{fig:reconstr_train_size} shows that the equivariant reconstruction error remains consistently low across models trained on 50\% (550K), 25\% (275K), 10\% (110K), and 1\% (11K) of the data (the full dataset contains 1.1M molecules), indicating that data augmentation alone is sufficient to induce equivariance even in low-data regimes.

\begin{figure}[htbp!]

\centering
\centerline{\includegraphics[width=\columnwidth]{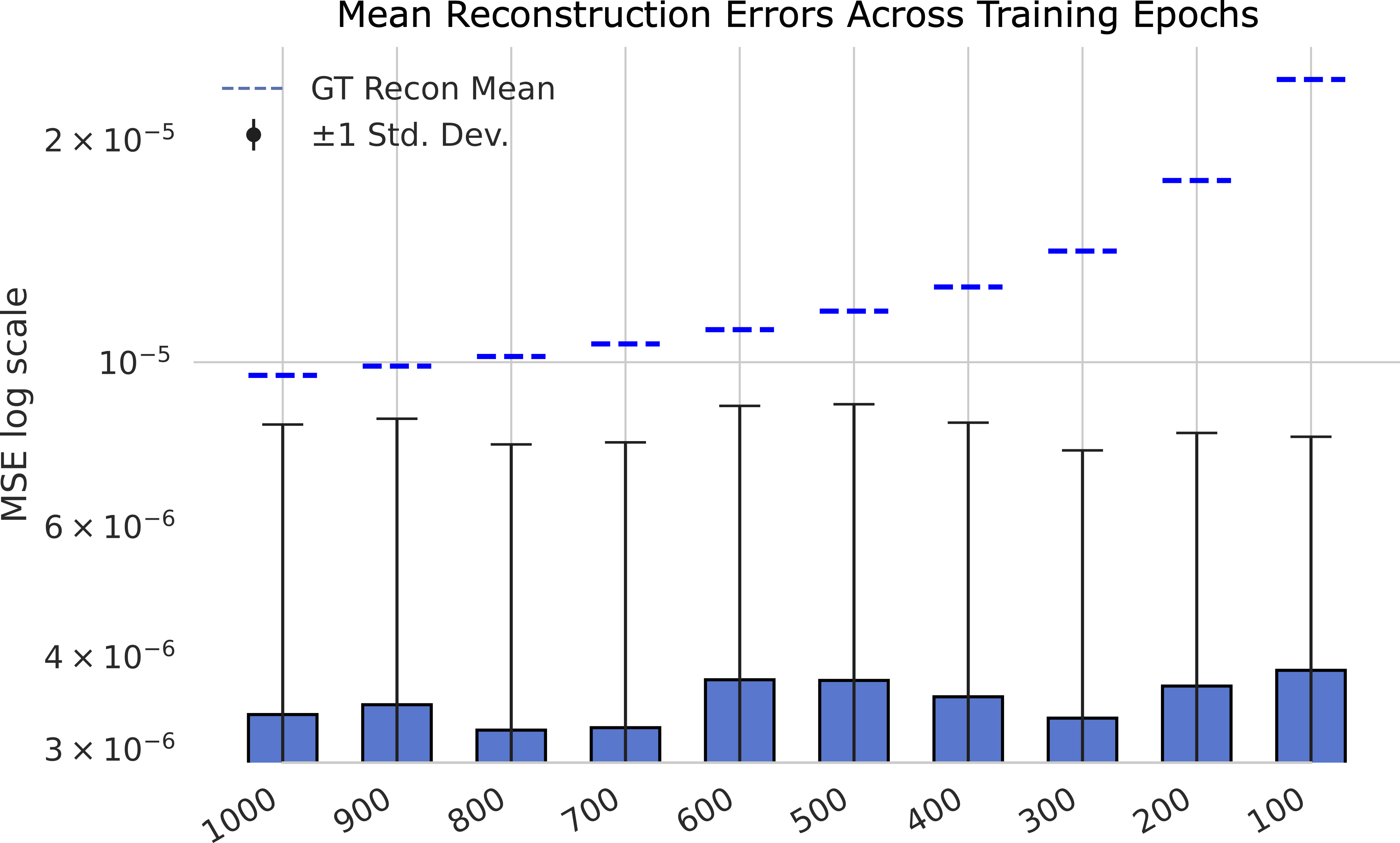}}
\caption{Effect of the \textbf{number of training epochs}. Equivariance is learned in very early epochs although the overall reconstruction errors are higher in earlier epochs.}
\label{fig:reconstr_epochs}
\vskip -0.2in
\end{figure}

\textbf{Effect of  Training Duration.} Similarly, we assess models trained for 100 to 1,000 epochs (on the 10\% training set subset) in Figure~\ref{fig:reconstr_epochs}. Equivariance is learned early during training, with marginal improvements from longer training. While training for more epochs improves the overall reconstruction quality and consequently the generation performance, they do not meaningfully affect reconstruction equivariance.
 
\begin{figure*}[htbp!]

\centering
\centerline{\includegraphics[width=\textwidth]{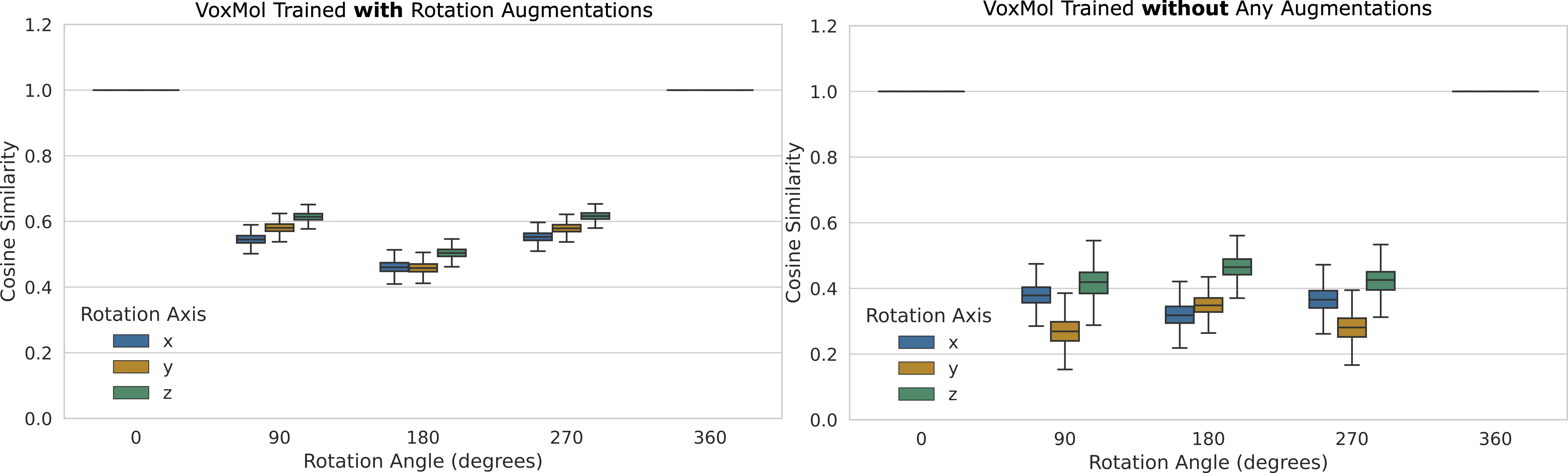}}
\caption{Cosine similarity between latent embeddings of the same molecules under different rotations. Latent embeddings are not rotation equivariant as rotated molecules are embedded differently. However, VoxMol trained with rotation augmentations embeds rotated molecules closer to each other compared to VoxMol trained with no augmentations, demostrating some learned equivariance in the latent space.}
\label{fig:reconstr_latents}
\vskip -0.2in
\end{figure*}

\textbf{Latent Embeddings}. Since the model is easily able to learn reconstruction equivariance without large training set sizes, with few training epochs, and even without explicit data augmentation, we seek to understand what feature of the training process leads to learned equivariance. Previous works have analyzed the latent embeddings of models to study the extent of their equivariance~\cite{kvinge2022ways}. An equivariant model should embed an input object to the same latent embeddings regardless of its rotation. 

We compute the cosine similarity between latent embeddings of the same molecule under different rotations and find that, except at 0\textdegree or 360\textdegree, VoxMol's embeddings differ significantly (see Figure~\ref{fig:reconstr_latents}). This indicates that the model does not recognize rotated molecules as the same object in latent space. 

However, we do observe that VoxMol trained with rotation augmentations embeds the rotated molecules closer together compared to VoxMol trained with no augmentations. In contrast, the explicitly equivariant E3NN model produces identical latent embeddings for rotated inputs, as expected (shown in Appendix Figure~\ref{fig:cos_e3nn}). 

These findings suggest that while VoxMol learns to reconstruct rotated inputs correctly, it does so by learning redundant latent representations. This inefficiency may require larger model capacity and could hinder tasks that rely on meaningful latent structure, such as property prediction or property guided generation~\cite{kaufman2024latent, xu2023geometric, huang2024learning}.

\subsection{Molecule Generation}

\begin{figure}[htbp!]
\centering
\centerline{\includegraphics[width=\columnwidth]{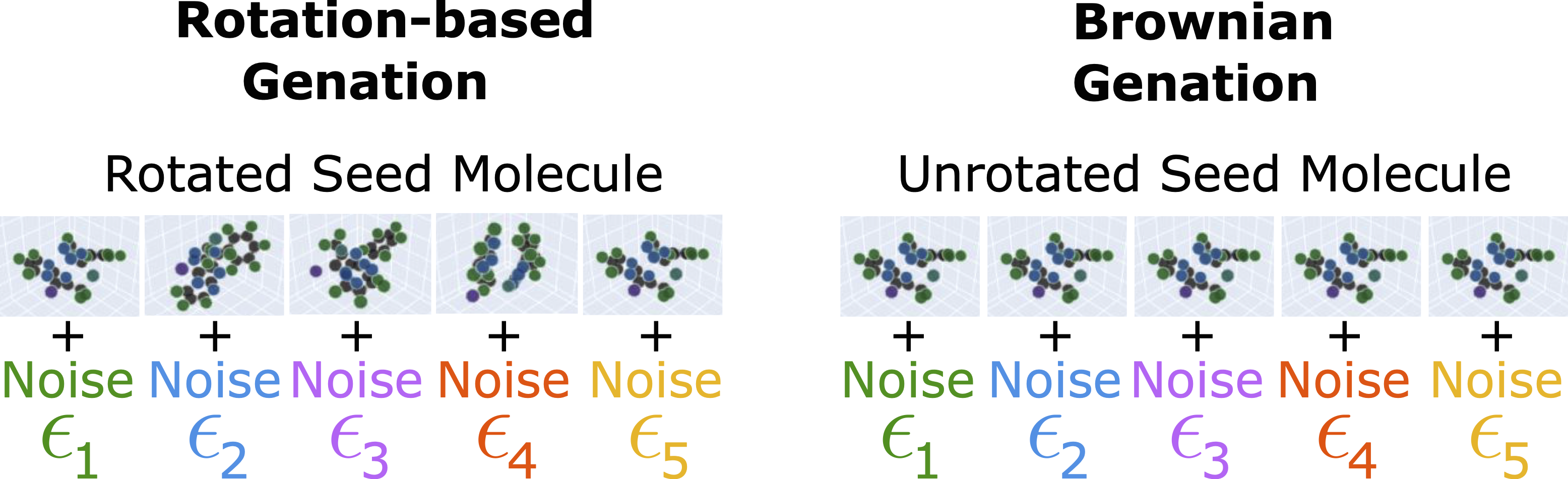}}
\caption{Experimental set up of generation equivariance. We compare generated molecules when the seed lead molecule is rotated to when the seed molecule is not rotated but noised with the same noise.}
\label{fig:gen_setup}
\end{figure}

\begin{figure}[htbp!]
\centering
\centerline{\includegraphics[width=\columnwidth]{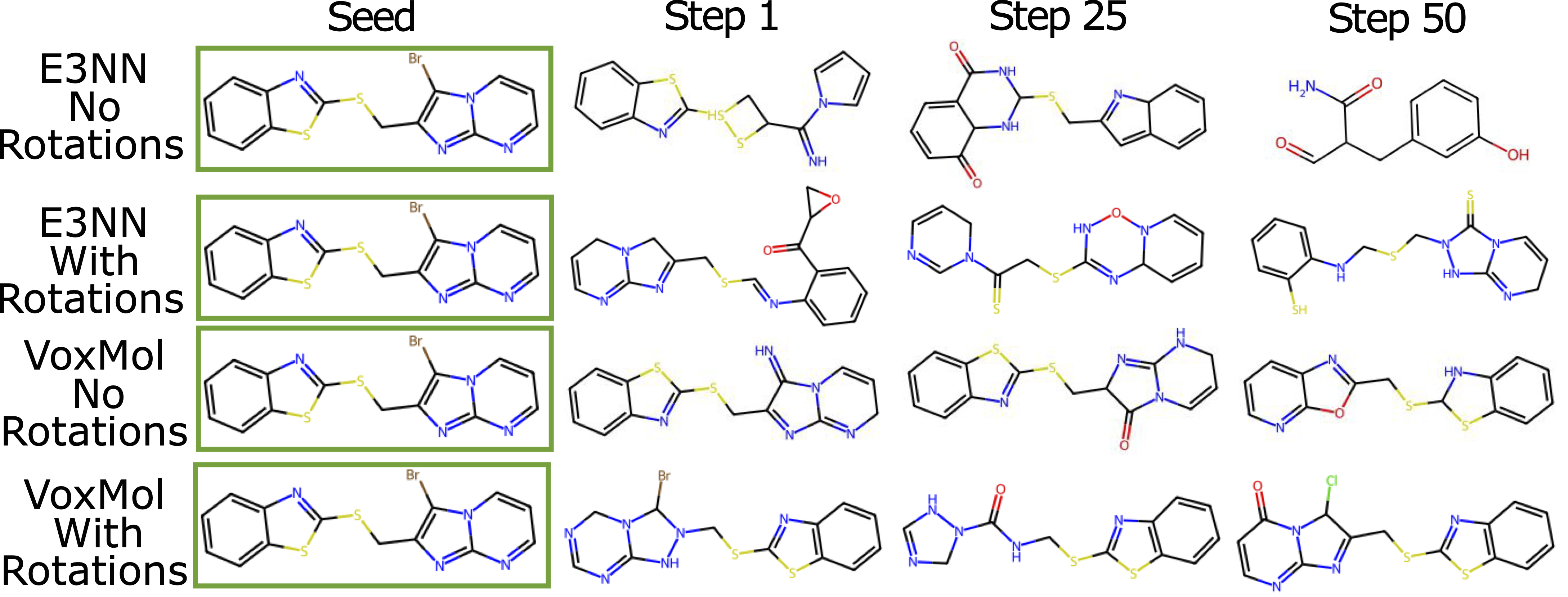}}
\caption{Generated molecules with and without rotating the seed molecule obtained with E3NN and VoxMol trained with rotations.}
\label{fig:gen_smiles}
\vskip -0.2in
\end{figure}

\begin{figure*}[htbp!]

\centering
\centerline{\includegraphics[width=\textwidth]{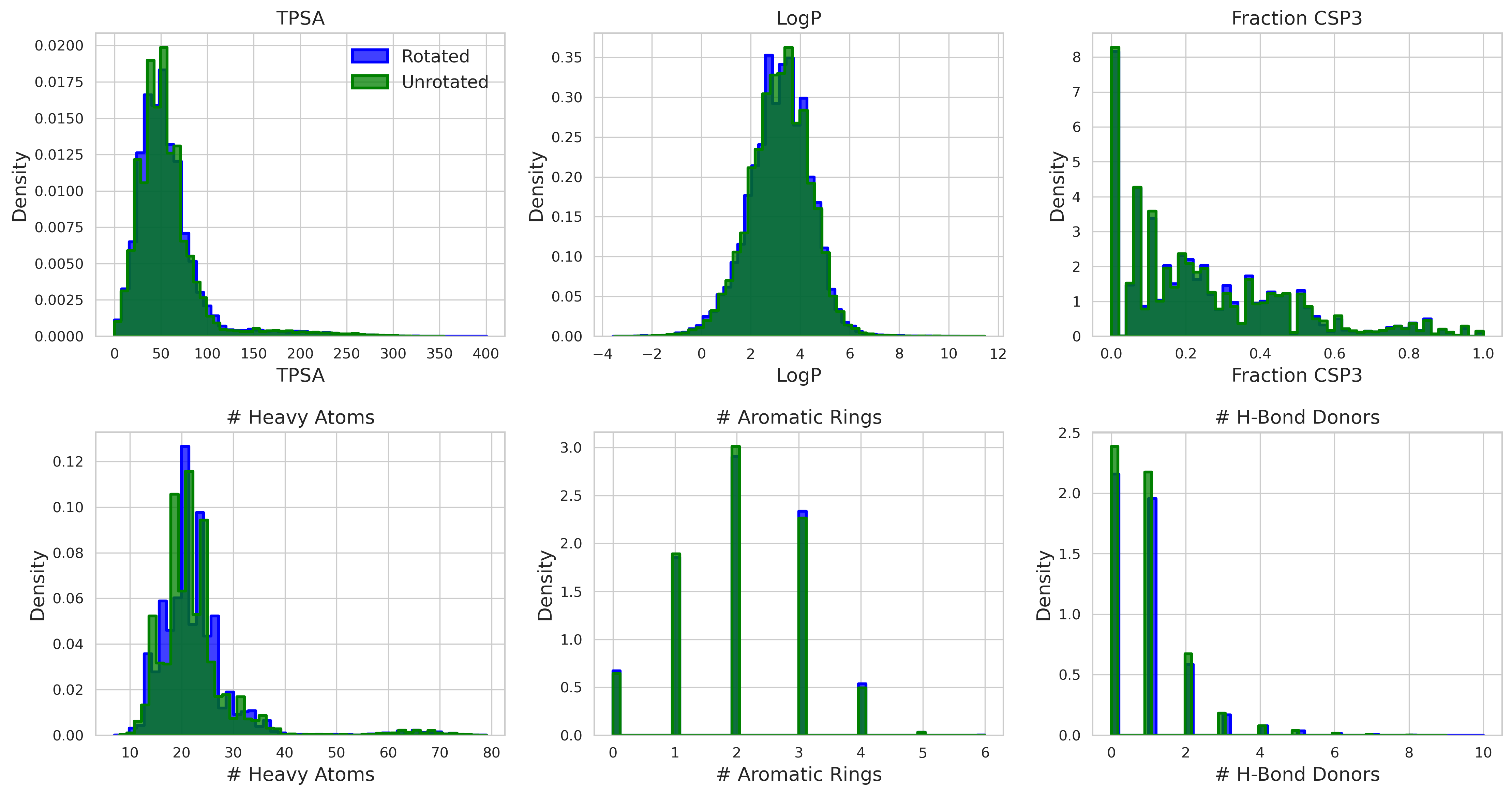}}
\caption{Distribution of chemical properties of generated molecules with and without rotating the seed molecule obtained with VoxMol trained with rotations.}
\label{fig:gen_props}
\vskip -0.2in
\end{figure*}

\begin{table*}[htbp!]
\centering
\resizebox{\textwidth}{!}{%
\begin{tabular}{l | x{25}x{25}x{25}x{25}x{25}x{25}}
  &  stable &  valid &  unique &  atom &  bond &  bond \\
  &  \%$  \uparrow$ & \%$ \uparrow$ & \%$ \uparrow$ &  TV$ \downarrow$ &  TV$ \downarrow$ &  ang\;W$ 1$$ \downarrow$ \\
E3NN - Unrotated Input & 41  \quad\tiny {($\pm$16)} & 85  \quad\tiny {($\pm$11)} & 95  \quad\tiny {($\pm$6)} & 0.130  \quad\tiny {($\pm$0.057)} & 0.054  \quad\tiny {($\pm$0.028)} & 3.983  \quad\tiny {($\pm$1.002)} \\
E3NN - Rotated Input & 41  \quad\tiny {($\pm$16)} & 86  \quad\tiny {($\pm$11)} & 95  \quad\tiny {($\pm$5)} & 0.129  \quad\tiny {($\pm$0.058)} & 0.054  \quad\tiny {($\pm$0.028)} & 3.987  \quad\tiny {($\pm$1.020)} \\
\hline
VoxMol - Unrotated Input & 72  \quad\tiny {($\pm$29)} & 81  \quad\tiny {($\pm$31)} & 57  \quad\tiny {($\pm$20)} & 0.188  \quad\tiny {($\pm$0.071)} & 0.058  \quad\tiny {($\pm$0.030)} & 3.567  \quad\tiny {($\pm$1.348)} \\
VoxMol - Rotated Input & 71  \quad\tiny {($\pm$29)} & 81  \quad\tiny {($\pm$31)} & 57  \quad\tiny {($\pm$20)} & 0.187  \quad\tiny {($\pm$0.071)} & 0.058  \quad\tiny {($\pm$0.030)} & \textbf{3.528}  \quad\tiny {($\pm$1.250)} \\
\hline
VoxMol (No Aug.) - Unrotated Input & \textbf{66}  \quad\tiny {($\pm$25)} & 84  \quad\tiny {($\pm$22)} & 75  \quad\tiny {($\pm$19)} & 0.187  \quad\tiny {($\pm$0.069)} & 0.057  \quad\tiny {($\pm$0.030)} & 3.600  \quad\tiny {($\pm$1.449)} \\
VoxMol (No Aug.) - Rotated Input & 61  \quad\tiny {($\pm$23)} & 85  \quad\tiny {($\pm$20)} & \textbf{83}  \quad\tiny {($\pm$17)} & \textbf{0.176}  \quad\tiny {($\pm$0.070)} & \textbf{0.056}  \quad\tiny {($\pm$0.029)} & \textbf{3.379}  \quad\tiny {($\pm$1.272)} \\
\hline
VoxMol (28 M) - Unrotated Input & \textbf{67}  \quad\tiny {($\pm$22)} & 84  \quad\tiny {($\pm$21)} & 77  \quad\tiny {($\pm$17)} & 0.169  \quad\tiny {($\pm$0.064)} & 0.057  \quad\tiny {($\pm$0.030)} & 3.244  \quad\tiny {($\pm$1.199)} \\
VoxMol (28 M) - Rotated Input & 64  \quad\tiny {($\pm$21)} & 84  \quad\tiny {($\pm$21)} & \textbf{81}  \quad\tiny {($\pm$15)} & \textbf{0.163}  \quad\tiny {($\pm$0.064)} & \textbf{0.055}  \quad\tiny {($\pm$0.029)} & \textbf{3.154}  \quad\tiny {($\pm$1.204)} \\

VoxMol (7 M) - Unrotated Input & \textbf{53} \quad\tiny{($\pm$18)}  & 84 \quad\tiny{($\pm$15)} & 90 \quad\tiny{($\pm$11)}  & 0.165 \quad\tiny{($\pm$0.060)} & 0.059 \quad\tiny{($\pm$0.030)}  & 3.469 \quad\tiny{($\pm$1.029)} \\

VoxMol (7 M) - Rotated Input & 49 \quad\tiny{($\pm$18)} & 84 \quad\tiny{($\pm$15)} & \textbf{92} \quad\tiny{($\pm$10)}  & 0.159 \quad\tiny{($\pm$0.059)} & 0.057 \quad\tiny{($\pm$0.029)} &  3.424 \quad\tiny{($\pm$0.989)} \\

\hline

VoxMol 50 \% Train Subset - Unrotated Input & \textbf{75} \quad\tiny{($\pm$30)} & 83 \quad\tiny{($\pm$30)} & 46 \quad\tiny{($\pm$20)} & 0.196 \quad\tiny{($\pm$0.074)} & 0.058 \quad\tiny{($\pm$0.031)} & 3.944 \quad\tiny{($\pm$1.677)} \\

VoxMol 50 \% Train Subset - Rotated Input & 70 \quad\tiny{($\pm$29)} & 84 \quad\tiny{($\pm$28)} & \textbf{53} \quad\tiny{($\pm$19)} &  \textbf{0.186} \quad\tiny{($\pm$0.075)} & \textbf{0.056} \quad\tiny{($\pm$0.030)}  & \textbf{3.531} \quad\tiny{($\pm$1.400)} \\

\end{tabular}
}
\caption{Quantitative metrics~\cite{vignac2023midi} are very similar for generations regardless whether the seed molecule is rotated or not with E3NN and VoxMol. But VoxMol models trained with no augmentations (``VoxMol (No Aug.)"), smaller models (``VoxMol (28M)" and ``VoxMol (7M)") or trained with less data (``VoxMol (50\% Train Subset)") exhibit lower molecular stability and higher uniqueness when inputs are rotated, suggesting they are not equivariant during generation and that their generations are lower quality.}
\label{tab:gen_midi}
\vskip -0.2in
\end{table*}

\begin{table*}[htbp!]
\centering
\resizebox{\textwidth}{!}{%
\begin{tabular}{lcccccc}
\toprule
\textbf{Model} & TPSA $ \downarrow$ & LogP $ \downarrow$ & Fraction CSP3 $ \downarrow$ & Heavy Atoms $ \downarrow$ & Aromatic Rings $ \downarrow$  & H-Bond Donors $ \downarrow$ \\
\midrule
E3NN & \textbf{0.0010} & \textbf{0.0013} & \textbf{0.0024} & \textbf{0.0023} & \textbf{0.0010} & \textbf{0.0001} \\
VoxMol (111 M) & \textbf{0.0025} & \textbf{0.0018} & \textbf{0.0029} & \textbf{0.0035} & \textbf{0.0007} & \textbf{0.0005} \\
VoxMol (No Aug.)  & 0.0222 & 0.0243 & 0.0203 & 0.0557 & 0.0104 & 0.0038 \\
VoxMol (28 M) & 0.0288 & 0.0118 & 0.0177 & 0.0308 & 0.0023 & 0.0037 \\
VoxMol (50 \%) & 0.0495 & 0.0492 & 0.0701 & 0.0729 & 0.0039 & 0.0127 \\

\bottomrule
\end{tabular}
}
\caption{KL Divergence between chemical property distributions of generated molecules with and without rotating the seed molecule. E3NN and VoxMol have the lowest KL divergence, showing that the generated distributions are very similar and not affected by rotations. On the other hand, VoxMol trained with no augmentations (``VoxMol (No Aug.)"), smaller model (``VoxMol (28M)") and with less training data (``VoxMol (50\% Train Subset)") generate more different distributions with higher KL divergence between generated properties.}
\label{tab:gen_kl}
\vskip -0.2in
\end{table*}

Reconstruction equivariance does not necessarily imply equivariance during generation nor does it guarantee high-quality generations. To evaluate how rotations affect the distribution of generated molecules, we conduct a series of generation experiments to assess whether models maintain equivariance beyond denoising. We perform two controlled experiments for each model. In the first experiment, we initialize generation with the same seed lead molecule rotated by 0\textdegree, 90\textdegree, 180\textdegree, 270\textdegree, and 360\textdegree, testing how rotation impacts the generations. In the second experiment, we fix the seed molecule's orientation and generate the same number of molecules as in the first experiment using the same noising for reproducibility, to test how randomness from ``Brownian motion" alone in WJS affects the generations compared to rotations of the seed molecule. Figure~\ref{fig:gen_setup} shows the experimental set up for generation equivariance. 

In both experiments, we run 5 chains for 100 denoising steps each, saving intermediate molecules every 10 steps, repeated for 100 seed ligands. This results in 5,000 generated molecules per experiment. We compare generations across four models: E3NN (equivariant by design), full-size VoxMol, VoxMol trained with no augmentations, a smaller VoxMol (with 7 M and 28M parameters), and VoxMol trained on only 50\% of the dataset.

\textbf{Generation Quality and Diversity.} To measure robustness to rotations, we compare commonly used generation quality metrics, such as stability, validity, uniqueness, and atom and bond distributions reported in Table~\ref{tab:gen_midi}. Generations with E3NN and full-size VoxMol are largely unaffected by seed molecule rotations, yielding similar stability and uniqueness metrics across experiments with and without rotating the seed molecule. In contrast, models trained with no augmentations (``VoxMol (No Aug.)"), smaller model capacity (``VoxMol (28M)" and ``VoxMol (7M)"), or reduced training data (``VoxMol (50\% Train Subset)") exhibit decreased stability and increased uniqueness under rotation. These results suggest that while VoxMol can learn equivariance during reconstruction, it may fail to generalize this robustness to generation without sufficient capacity and data augmentation. Figure~\ref{fig:gen_smiles} shows example generated molecules.

\textbf{Property Distribution Shifts Under Rotation.} We also computed chemical properties of generated molecules using RDKit~\cite{landrum2016rdkit} and computed the Kullback–Leibler (KL) divergence between the distributions of each property generated with and without rotations of the seed molecule. As seen in Table~\ref{tab:gen_kl} E3NN and VoxMol show low divergence between rotated and unrotated generations, indicating their generations are very similar across orientations. However, models trained with no augmentation, smaller model size, and limited data all show higher KL divergence, revealing that rotations meaningfully alter their generation behavior. This suggests that these models are not fully equivariant during generation. We visualize histograms of properties of generated molecules with and without rotating the seed molecule obtained with VoxMol in Figure~\ref{fig:gen_props}.

\subsection{Property Predictions}

\begin{table*}[htbp!]
\centering
\resizebox{\textwidth}{!}{%
\begin{tabular}{llcccc}
\toprule
\textbf{Model} & \textbf{Setup} & \textbf{Spearman (GT)} $ \uparrow$ & \textbf{Spearman (Rot)} $ \uparrow$ & \textbf{MAE (GT)} $ \downarrow$ & \textbf{MAE (Rot)} $ \downarrow$ \\
\midrule

VoxMol (111 M)      & Encoder-only      & 0.83 & 0.95 & 0.054 & 0.021 \\
VoxMol (111 M)      & Enc+Dec+Denoise   & 0.84 & 0.96 & 0.046 & 0.019 \\
VoxMol (No Aug.)    & Encoder-only      & 0.77 & 0.43 & 0.068 & 0.150 \\
VoxMol (No Aug.)    & Enc+Dec+Denoise   & 0.89 & 0.64 & 0.035 & 0.130 \\
VoxMol (7 M)        & Encoder-only      & 0.84 & 0.93 & 0.050 & 0.028 \\
VoxMol (7 M)        & Enc+Dec+Denoise   & 0.84 & 0.94 & 0.047 & 0.025 \\

\bottomrule
\end{tabular}
}
\caption{Property (asphericity) prediction equivariance under input rotations. Spearman correlation and Mean Absolute Error (MAE) are computed between predicted and ground truth values (Spearman-GT, MAE-GT) and between predictions on rotated versions of the same molecule (Spearman-Rot, MAE-Rot).}
\label{tab:pred}
\vskip -0.2in
\end{table*}

To evaluate how input rotations affect property prediction, we train each model to predict asphericity, a geometry-sensitive property that measures how close a 3D object is to a perfect sphere. We add a predictive head to each model and compare the full-size VoxMol model (architecture details are provided in the Appendix), trained without rotational augmentations and a smaller model. For evaluation, we compute Spearman correlation and Mean Absolute Error (MAE) between predicted and ground truth values (Spearman-GT, MAE-GT). To test rotation robustness, we compute the same metrics between predictions on rotated versions of the same molecule (Spearman-Rot, MAE-Rot). If a model is equivariant, predictions should remain consistent across rotations~\cite{gerken2024emergent}.

We compare training with encoders only, where each model is trained to predict just the property (``Encoder-only"), and where each model is jointly trained to reconstruct the input molecule with an additional MSE voxel denoising loss used in addition to predicting the property (``Enc+Dec+Denoise"). All inputs are noised with the same noise as during training for reconstruction and during generation to better analyze the effect of the additional reconstruction (denoising) loss. Our results show that adding a reconstruction loss consistently improves both prediction accuracy and rotation robustness, suggesting it encourages the model to encode more spatially meaningful features (see Table~\ref{tab:pred}). VoxMol trained without augmentation performs the worst and is clearly non-equivariant. Interestingly, a smaller VoxMol 7M model performs nearly as well as the full 111M parameter model, highlighting that model size is not the limiting factor for learning equivariant predictions unlike during generation.

\section{Conclusions}
\label{conclusions}

Our results show that CNNs trained with rotation augmentation easily learn equivariance for reconstruction of noised molecules, even with small models, limited data and few training epochs. However, achieving robustness to rotations during generation requires larger models, larger datasets, and longer training. 


Since generative models are trained only with a simple reconstruction loss which saturates easily, it may not fully allow the model to learn features useful for good performance on generation and prediction tasks. Therefore, large models and training datasets may be required because the model may need to learn redundant features. This is illustrated by very different latent embeddings of the same molecule at different rotations. 

The training process could be improved with additional auxiliary tasks and regularization that would encourage the model to learn similar latent embeddings for the same molecules under different rotations. This may not only improve the model's equivariance during generation and prediction but perhaps could also improve the quality of generations by adding inductive bias and reduce the need for very large models that require long training times and large datasets due to their large number of parameters.

\section{Impact Statement}
This work aims to advance Machine Learning by studying learned equivariance in molecular generative models. Our findings may simplify model design and improve scalability for applications in drug discovery and materials science. We do not anticipate any specific ethical or societal risks.

\clearpage
\bibliography{example_paper}

\begin{thebibliography}{27}
\providecommand{\natexlab}[1]{#1}
\providecommand{\url}[1]{\texttt{#1}}
\expandafter\ifx\csname urlstyle\endcsname\relax
  \providecommand{\doi}[1]{doi: #1}\else
  \providecommand{\doi}{doi: \begingroup \urlstyle{rm}\Url}\fi

\bibitem[Axelrod \& Gomez-Bombarelli(2022)Axelrod and
  Gomez-Bombarelli]{geomdrugs_axelrod2022geom}
Axelrod, S. and Gomez-Bombarelli, R.
\newblock Geom, energy-annotated molecular conformations for property
  prediction and molecular generation.
\newblock \emph{Scientific Data}, 9\penalty0 (1):\penalty0 185, 2022.

\bibitem[Bao et~al.(2022)Bao, Zhao, Hao, Li, Li, and Zhu]{EEG-SDE}
Bao, F., Zhao, M., Hao, Z., Li, P., Li, C., and Zhu, J.
\newblock Equivariant energy-guided sde for inverse molecular design.
\newblock \emph{arXiv preprint arXiv:2209.15408}, 2022.

\bibitem[Bronstein et~al.(2021)Bronstein, Bruna, Cohen, and
  Veli{\v{c}}kovi{\'c}]{bronstein2021geometric}
Bronstein, M.~M., Bruna, J., Cohen, T., and Veli{\v{c}}kovi{\'c}, P.
\newblock Geometric deep learning: Grids, groups, graphs, geodesics, and
  gauges.
\newblock \emph{arXiv preprint arXiv:2104.13478}, 2021.

\bibitem[Cheng et~al.(2018)Cheng, Chatterji, Bartlett, and
  Jordan]{cheng2018underdamped}
Cheng, X., Chatterji, N.~S., Bartlett, P.~L., and Jordan, M.~I.
\newblock Underdamped langevin mcmc: A non-asymptotic analysis.
\newblock In \emph{Conference on learning theory}, pp.\  300--323. PMLR, 2018.

\bibitem[Cui et~al.(2023)Cui, Mittal, Lu, Zhang, Saon, and
  Kingsbury]{cui2023soft}
Cui, X., Mittal, A., Lu, S., Zhang, W., Saon, G., and Kingsbury, B.
\newblock Soft random sampling: A theoretical and empirical analysis.
\newblock \emph{arXiv preprint arXiv:2311.12727}, 2023.

\bibitem[Diaz et~al.(2023)Diaz, Geiger, and McKinley]{diaz2023end}
Diaz, I., Geiger, M., and McKinley, R.~I.
\newblock An end-to-end se (3)-equivariant segmentation network.
\newblock \emph{arXiv preprint arXiv:2303.00351}, 2023.

\bibitem[Fei \& Deng(2024)Fei and Deng]{fei2024rotation}
Fei, J. and Deng, Z.
\newblock Rotation invariance and equivariance in 3d deep learning: a survey.
\newblock \emph{Artificial Intelligence Review}, 57\penalty0 (7):\penalty0 168,
  2024.

\bibitem[Gebauer et~al.(2019)Gebauer, Gastegger, and
  Sch{\"u}tt]{gebauer2019symmetry}
Gebauer, N., Gastegger, M., and Sch{\"u}tt, K.
\newblock Symmetry-adapted generation of 3d point sets for the targeted
  discovery of molecules.
\newblock \emph{Advances in neural information processing systems}, 32, 2019.

\bibitem[Gerken et~al.(2022)Gerken, Carlsson, Linander, Ohlsson, Petersson, and
  Persson]{gerken2022equivariance}
Gerken, J., Carlsson, O., Linander, H., Ohlsson, F., Petersson, C., and
  Persson, D.
\newblock Equivariance versus augmentation for spherical images.
\newblock In \emph{International Conference on Machine Learning}, pp.\
  7404--7421. PMLR, 2022.

\bibitem[Gerken \& Kessel(2024)Gerken and Kessel]{gerken2024emergent}
Gerken, J.~E. and Kessel, P.
\newblock Emergent equivariance in deep ensembles.
\newblock \emph{arXiv preprint arXiv:2403.03103}, 2024.

\bibitem[Hoogeboom et~al.(2022)Hoogeboom, Satorras, Vignac, and
  Welling]{hoogeboom2022equivariant}
Hoogeboom, E., Satorras, V.~G., Vignac, C., and Welling, M.
\newblock Equivariant diffusion for molecule generation in 3d.
\newblock In \emph{International conference on machine learning}, pp.\
  8867--8887, 2022.

\bibitem[Huang et~al.(2024)Huang, Sun, Du, and Lv]{huang2024learning}
Huang, H., Sun, L., Du, B., and Lv, W.
\newblock Learning joint 2-d and 3-d graph diffusion models for complete
  molecule generation.
\newblock \emph{IEEE Transactions on Neural Networks and Learning Systems},
  2024.

\bibitem[Kaufman et~al.(2024)Kaufman, Williams, Pederson, Underkoffler,
  Panjwani, Wang-Henderson, Mardirossian, Katcher, Strater, Grandjean,
  et~al.]{kaufman2024latent}
Kaufman, B., Williams, E.~C., Pederson, R., Underkoffler, C., Panjwani, Z.,
  Wang-Henderson, M., Mardirossian, N., Katcher, M.~H., Strater, Z., Grandjean,
  J.-M., et~al.
\newblock Latent diffusion for conditional generation of molecules.
\newblock \emph{bioRxiv}, pp.\  2024--08, 2024.

\bibitem[Kingma \& Welling(2014)Kingma and Welling]{kingma2013auto}
Kingma, D.~P. and Welling, M.
\newblock Auto-encoding variational bayes.
\newblock \emph{International Conference on Learning Representations (ICLR)},
  2014.

\bibitem[Kvinge et~al.(2022)Kvinge, Emerson, Jorgenson, Vasquez, Doster, and
  Lew]{kvinge2022ways}
Kvinge, H., Emerson, T., Jorgenson, G., Vasquez, S., Doster, T., and Lew, J.
\newblock In what ways are deep neural networks invariant and how should we
  measure this?
\newblock \emph{Advances in Neural Information Processing Systems},
  35:\penalty0 32816--32829, 2022.

\bibitem[Landrum et~al.(2016)]{landrum2016rdkit}
Landrum, G. et~al.
\newblock Rdkit: Open-source cheminformatics software. 2016.
\newblock \emph{URL http://www. rdkit. org/, https://github. com/rdkit/rdkit},
  149\penalty0 (150):\penalty0 650, 2016.

\bibitem[Nowara et~al.(2024)Nowara, Pinheiro, Mahajan, Mahmood, Watkins,
  Saremi, and Maser]{nowara2024nebula}
Nowara, E., Pinheiro, P.~O., Mahajan, S.~P., Mahmood, O., Watkins, A.~M.,
  Saremi, S., and Maser, M.
\newblock Nebula: Neural empirical bayes under latent representations for
  efficient and controllable design of molecular libraries.
\newblock In \emph{ICML 2024 AI for Science Workshop}, 2024.

\bibitem[Pinheiro et~al.(2023)Pinheiro, Rackers, Kleinhenz, Maser, Mahmood,
  Watkins, Ra, Sresht, and Saremi]{pinheiro20233d}
Pinheiro, P.~O., Rackers, J., Kleinhenz, J., Maser, M., Mahmood, O., Watkins,
  A.~M., Ra, S., Sresht, V., and Saremi, S.
\newblock 3d molecule generation by denoising voxel grids.
\newblock \emph{Advances in Neural Information Processing Systems}, 2023.

\bibitem[Pinheiro et~al.(2024)Pinheiro, Jamasb, Mahmood, Sresht, and
  Saremi]{pinheiro2024structure}
Pinheiro, P.~O., Jamasb, A., Mahmood, O., Sresht, V., and Saremi, S.
\newblock Structure-based drug design by denoising voxel grids.
\newblock \emph{arXiv preprint arXiv:2405.03961}, 2024.

\bibitem[Quiroga et~al.(2020)Quiroga, Ronchetti, Lanzarini, and
  Bariviera]{quiroga2020revisiting}
Quiroga, F., Ronchetti, F., Lanzarini, L., and Bariviera, A.~F.
\newblock Revisiting data augmentation for rotational invariance in
  convolutional neural networks.
\newblock In \emph{Modelling and Simulation in Management Sciences: Proceedings
  of the International Conference on Modelling and Simulation in Management
  Sciences (MS-18)}, pp.\  127--141. Springer, 2020.

\bibitem[Ragoza et~al.(2020)Ragoza, Masuda, and Koes]{ragoza2020learning}
Ragoza, M., Masuda, T., and Koes, D.~R.
\newblock Learning a continuous representation of 3d molecular structures with
  deep generative models.
\newblock \emph{arXiv preprint arXiv:2010.08687}, 2020.

\bibitem[Rombach et~al.(2022)Rombach, Blattmann, Lorenz, Esser, and
  Ommer]{rombach2022high}
Rombach, R., Blattmann, A., Lorenz, D., Esser, P., and Ommer, B.
\newblock High-resolution image synthesis with latent diffusion models.
\newblock In \emph{Proceedings of the IEEE/CVF conference on computer vision
  and pattern recognition}, pp.\  10684--10695, 2022.

\bibitem[Saremi \& Hyv{\"a}rinen(2019)Saremi and
  Hyv{\"a}rinen]{saremi2019neural}
Saremi, S. and Hyv{\"a}rinen, A.
\newblock Neural empirical bayes.
\newblock \emph{Journal of Machine Learning Research}, 20\penalty0
  (181):\penalty0 1--23, 2019.

\bibitem[Skalic et~al.(2019)Skalic, Jim{\'e}nez, Sabbadin, and
  De~Fabritiis]{skalic2019shape}
Skalic, M., Jim{\'e}nez, J., Sabbadin, D., and De~Fabritiis, G.
\newblock Shape-based generative modeling for de novo drug design.
\newblock \emph{Journal of chemical information and modeling}, 59\penalty0
  (3):\penalty0 1205--1214, 2019.

\bibitem[Vignac et~al.(2023)Vignac, Osman, Toni, and Frossard]{vignac2023midi}
Vignac, C., Osman, N., Toni, L., and Frossard, P.
\newblock Midi: Mixed graph and 3d denoising diffusion for molecule generation.
\newblock \emph{arXiv preprint arXiv:2302.09048}, 2023.

\bibitem[Weiler et~al.(2018)Weiler, Geiger, Welling, Boomsma, and
  Cohen]{weiler20183d}
Weiler, M., Geiger, M., Welling, M., Boomsma, W., and Cohen, T.~S.
\newblock 3d steerable cnns: Learning rotationally equivariant features in
  volumetric data.
\newblock \emph{Advances in Neural Information Processing Systems}, 31, 2018.

\bibitem[Xu et~al.(2023)Xu, Powers, Dror, Ermon, and Leskovec]{xu2023geometric}
Xu, M., Powers, A.~S., Dror, R.~O., Ermon, S., and Leskovec, J.
\newblock Geometric latent diffusion models for 3d molecule generation.
\newblock In \emph{International Conference on Machine Learning}, pp.\
  38592--38610. PMLR, 2023.

\end{thebibliography}
\bibliographystyle{icml2025}

\newpage
\appendix
\onecolumn

\clearpage
\FloatBarrier

\section{Appendix}

\subsection{Model Architecture}
\label{sec:appendix_arch}

We train two models to evaluate learned equivariance. 

\textbf{VoxMol} is based on a U-Net model with 3D CNN layers with 4 levels of resolution and self-attention on the lowest two resolutions~\cite{pinheiro20233d}. VoxMol was trained until convergence for 239 epochs with noise level of $\sigma = 0.9$ and learning rate of 1e-5. We trained VoxMol with data augmentation by randomly rotating and translating every sample in each iteration. See~\citet{pinheiro20233d} for architecture and training details. 

\textbf{E3NN model} is an equivariant version of VoxMol with E3NN layers. We use the SE(3)-equivariant 3D U-Net using steerable
CNNs~\cite{weiler20183d} and we use the official implementation
of~\citet{diaz2023end}. We tune the network hyperparameters for the best denoising performance, however, the E3NN model is not able to match the performance of VoxMol on denoising or generation metrics. We trained E3NN until convergence for 138 epochs and learning rate of 1e-2.

\textbf{Predictive Head}
We add a lightweight Multi-Layer Perceptron (MLP) head to the UNet encoder for property prediction. The MLP includes average pooling to reduce spatial dimensions, followed by three linear layers with layer normalization, dropout of 0.1, and Shifted SoftPlus activations. We use Tanhshrink as the activation in the final layer for the regression task. The input voxels were noised with $\sigma=0.9$ as in the reconstruction task.

\subsection{Dataset}
\label{sec:appendix_dataset}

We train all models on a standard public dataset used for molecule generation called GEOM-drugs (GEOM)~\cite{geomdrugs_axelrod2022geom}, following the train, validation, and test set splits used in~\cite{vignac2023midi} with 1.1M/146K/146K molecules each. We train every model with soft random subsampling~\cite{cui2023soft} and train on 10\% of the training set in each epoch. We use voxel grids of dimension 64, and 8 atom channels ([C, H, O, N, F, S, Cl, Br]) with atomic radii of 0.25 $\AA$ resolution.

\subsection{Additional Reconstruction Results}
\label{sec:appendix_rec}

\begin{figure}[htbp!]

\centering
\centerline{\includegraphics[width=\columnwidth]{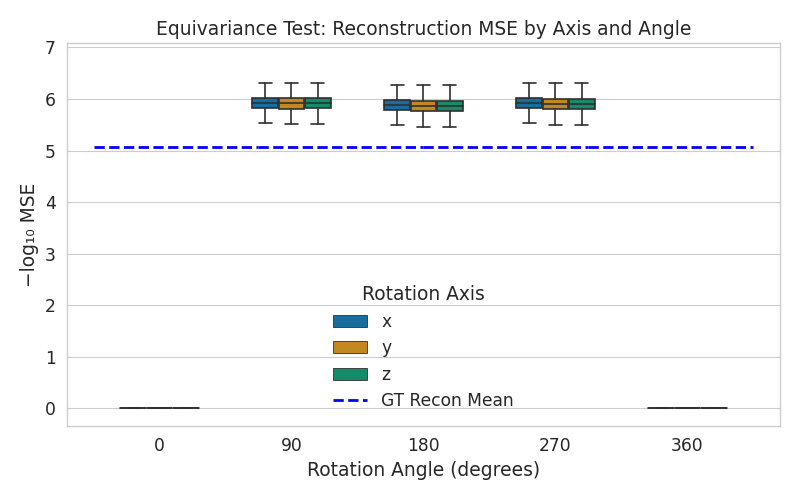}}
\caption{Rotational equivariance for the \textbf{111 M model}. }
\label{fig:reconstr_vox}
\vskip -0.2in
\end{figure}

\begin{figure}[htbp!]

\centering
\centerline{\includegraphics[width=\columnwidth]{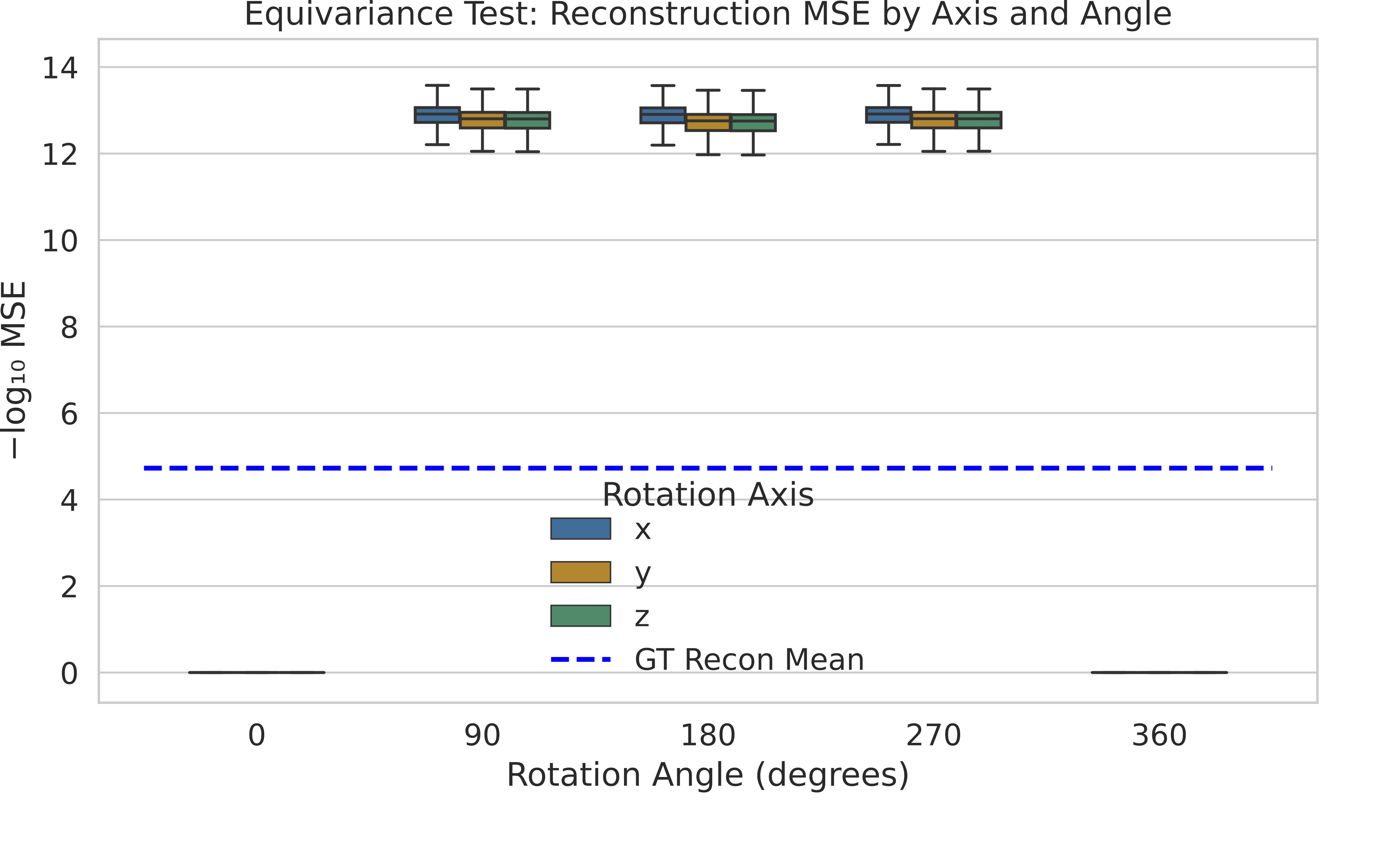}}
\caption{Rotational equivariance for the \textbf{E3NN model}. }
\label{fig:reconstr_e3nn}
\vskip -0.2in
\end{figure}

\begin{figure}[htbp!]

\centering
\centerline{\includegraphics[width=\columnwidth]{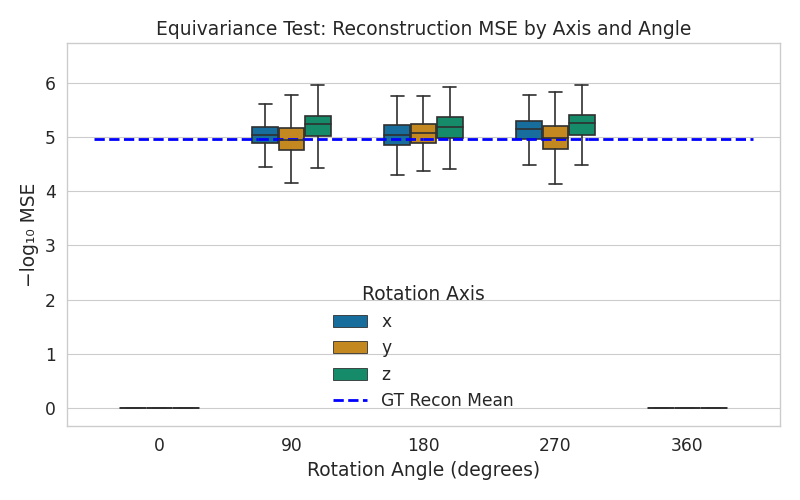}}
\caption{Rotational equivariance for the 111 M trained with \textbf{no rotation augmentations}.}
\label{fig:reconstr_no_rot}
\vskip -0.2in
\end{figure}

\begin{figure}[htbp!]

\centering
\centerline{\includegraphics[width=\columnwidth]{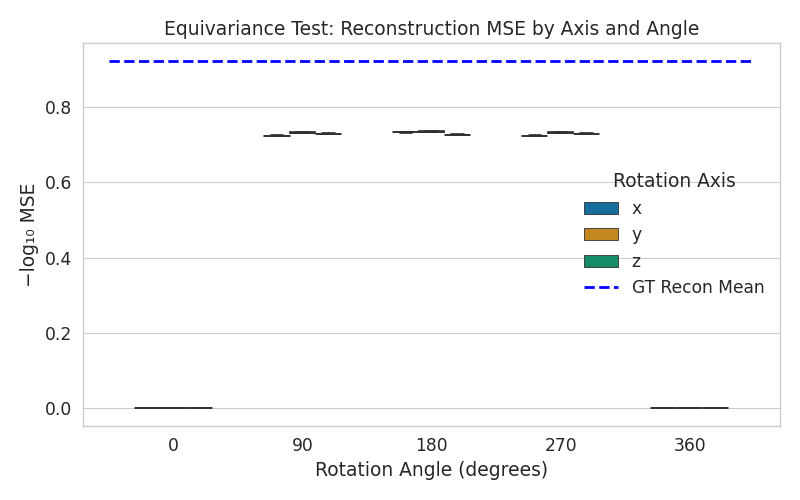}}
\caption{Rotational equivariance for the 111 M model with \textbf{randomly initialized weights} before training.}
\label{fig:reconstr_rand}
\vskip -0.2in
\end{figure}

\begin{figure}[htbp!]

\centering
\centerline{\includegraphics[width=\columnwidth]{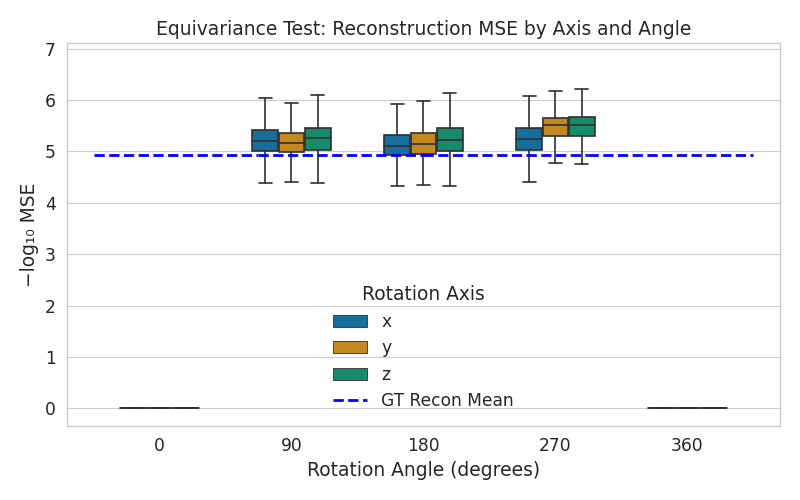}}
\caption{Rotational equivariance for the \textbf{28 M model}.}
\label{fig:reconstr_small}
\vskip -0.2in
\end{figure}

\begin{figure}[htbp!]

\centering
\centerline{\includegraphics[width=\columnwidth]{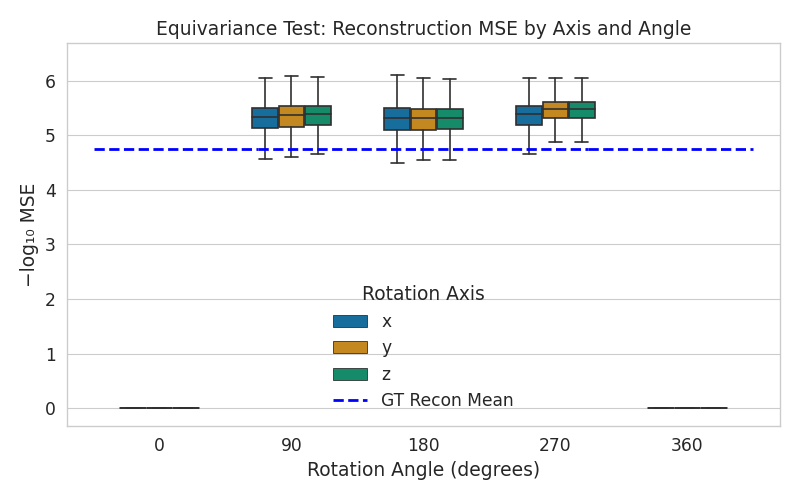}}
\caption{Rotational equivariance for the \textbf{7 M model}.}
\label{fig:reconstr_tiny}
\vskip -0.2in
\end{figure}

%

\begin{figure}[htbp!]

\centering
\centerline{\includegraphics[width=\columnwidth]{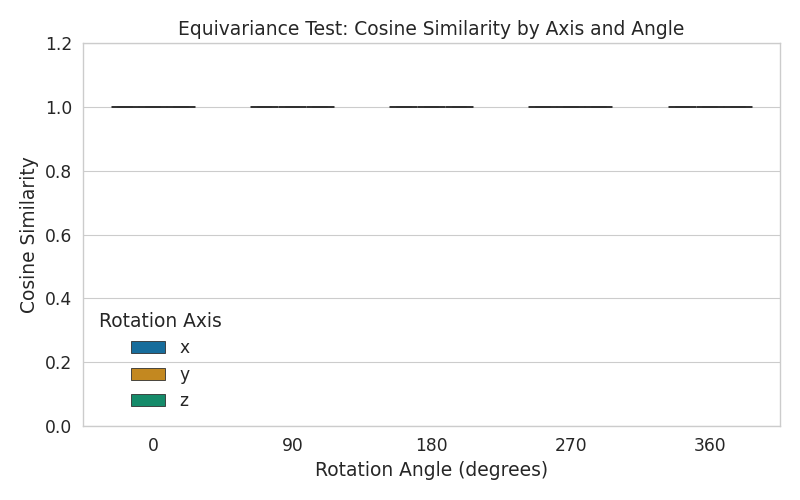}}
\caption{Cosine similarity between latent embeddings of the same molecules under different rotations for the \textbf{E3NN model}.}
\label{fig:cos_e3nn}
\vskip -0.2in
\end{figure}

%

\begin{figure}[htbp!]

\centering
\centerline{\includegraphics[width=\columnwidth]{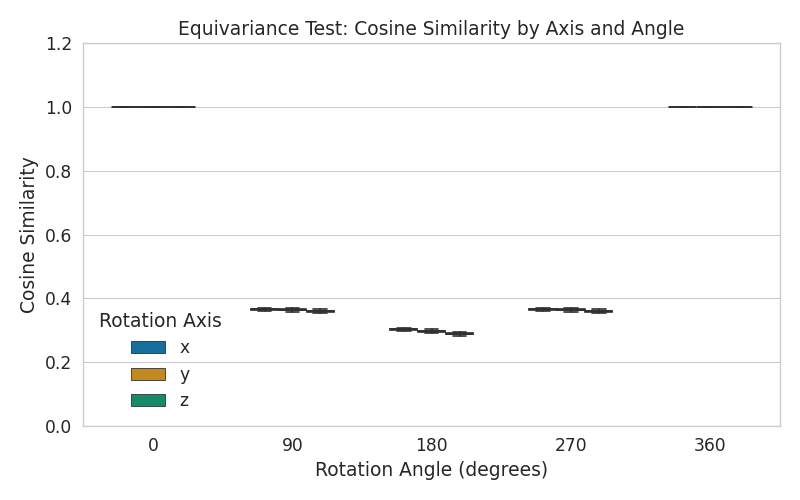}}
\caption{Cosine similarity between latent embeddings of the same molecules under different rotations for the 111 M model with \textbf{randomly initialized weights} before training.}
\label{fig:cos_rand}
\vskip -0.2in
\end{figure}

\begin{figure}[htbp!]

\centering
\centerline{\includegraphics[width=\columnwidth]{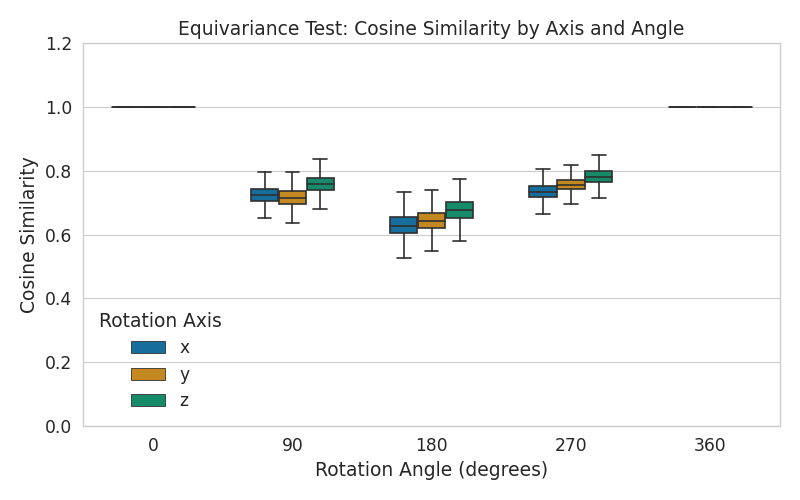}}
\caption{Cosine similarity between latent embeddings of the same molecules under different rotations for the \textbf{28 M model}.}
\label{fig:cos_small}
\vskip -0.2in
\end{figure}

\begin{figure}[htbp!]

\centering
\centerline{\includegraphics[width=\columnwidth]{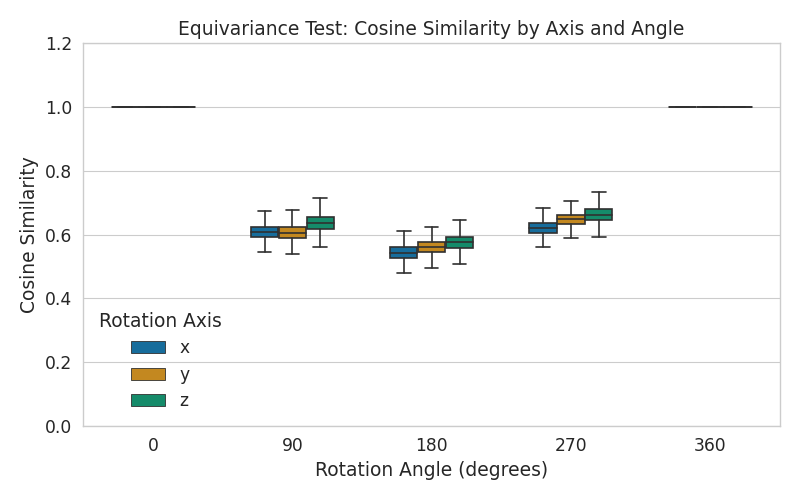}}
\caption{Cosine similarity between latent embeddings of the same molecules under different rotations for the \textbf{7 M model}.}
\label{fig:cos_tiny}
\vskip -0.2in
\end{figure}

\clearpage

\subsection{Additional Generation Results}

\begin{table*}
\begin{tabular}{l | x{25}x{25}x{25}x{25}x{25}x{25}x{25}x{25}x{25}}
  & \tiny stable & \tiny stable & \tiny valid & \tiny unique & \tiny valency & \tiny atom & \tiny bond & \tiny bond & \tiny bond \\
  & \tiny mol. $ \uparrow$ & \tiny atom $ \uparrow$ & \tiny\%$ \uparrow$ & \tiny\%$ \uparrow$ & \tiny W$ 1$$ \downarrow$ & \tiny TV$ \downarrow$ & \tiny TV$ \downarrow$ & \tiny len\;W$ 1$$ \downarrow$ & \tiny ang\;W$ 1$$ \downarrow$ \\
E3NN - Unrotated Input & 41 \quad\tiny{($\pm$16)} & 97 \quad\tiny{($\pm$1)} & 85 \quad\tiny{($\pm$11)} & 95 \quad\tiny{($\pm$6)} & 28 \quad\tiny{($\pm$2)} & 0.130 \quad\tiny{($\pm$0.057)} & 0.054 \quad\tiny{($\pm$0.028)} & 0.004 \quad\tiny{($\pm$0.001)} & 3.983 \quad\tiny{($\pm$1.002)} \\
E3NN - Rotated Input & 41 \quad\tiny{($\pm$16)} & 97 \quad\tiny{($\pm$1)} & 86 \quad\tiny{($\pm$11)} & 95 \quad\tiny{($\pm$5)} & 28 \quad\tiny{($\pm$2)} & 0.129 \quad\tiny{($\pm$0.058)} & 0.054 \quad\tiny{($\pm$0.028)} & 0.004 \quad\tiny{($\pm$0.001)} & 3.987 \quad\tiny{($\pm$1.020)} \\
\hline

VoxMol - Unrotated Input & 72 \quad\tiny{($\pm$29)} & 99 \quad\tiny{($\pm$1)} & 81 \quad\tiny{($\pm$31)} & 57 \quad\tiny{($\pm$20)} & 28 \quad\tiny{($\pm$3)} & 0.188 \quad\tiny{($\pm$0.071)} & 0.058 \quad\tiny{($\pm$0.030)} & 0.005 \quad\tiny{($\pm$0.001)} & 3.567 \quad\tiny{($\pm$1.348)} \\
VoxMol - Rotated Input & 71 \quad\tiny{($\pm$29)} & 99 \quad\tiny{($\pm$1)} & 81 \quad\tiny{($\pm$31)} & 57 \quad\tiny{($\pm$20)} & 28 \quad\tiny{($\pm$3)} & 0.187 \quad\tiny{($\pm$0.071)} & 0.058 \quad\tiny{($\pm$0.030)} & 0.005 \quad\tiny{($\pm$0.001)} & 3.528 \quad\tiny{($\pm$1.250)} \\
\hline

VoxMol 111 M No Aug. - Unrotated Input & 66 \quad\tiny{($\pm$25)} & 99 \quad\tiny{($\pm$1)} & 84 \quad\tiny{($\pm$22)} & 75 \quad\tiny{($\pm$19)} & 26 \quad\tiny{($\pm$3)} & 0.187 \quad\tiny{($\pm$0.069)} & 0.057 \quad\tiny{($\pm$0.030)} & 0.004 \quad\tiny{($\pm$0.001)} & 3.600 \quad\tiny{($\pm$1.449)} \\
VoxMol 111 M No Aug. - Rotated Input & 61 \quad\tiny{($\pm$23)} & 99 \quad\tiny{($\pm$1)} & 85 \quad\tiny{($\pm$20)} & 83 \quad\tiny{($\pm$17)} & 25 \quad\tiny{($\pm$3)} & 0.176 \quad\tiny{($\pm$0.070)} & 0.056 \quad\tiny{($\pm$0.029)} & 0.004 \quad\tiny{($\pm$0.001)} & 3.379 \quad\tiny{($\pm$1.272)} \\

\end{tabular}
\caption{Detailed seeded Generation Results with models initialized with lead molecules without rotations (Brownian Motion (``BM")) and with rotations (``rot.").}
\label{tab:gen midi overlap1}
\vskip -0.2in
\end{table*}

\begin{table*}
\begin{tabular}{l | x{25}x{25}x{25}x{25}x{25}x{25}x{25}x{25}x{25}}
  & \tiny stable & \tiny stable & \tiny valid & \tiny unique & \tiny valency & \tiny atom & \tiny bond & \tiny bond & \tiny bond \\
  & \tiny mol. $ \uparrow$ & \tiny atom $ \uparrow$ & \tiny\%$ \uparrow$ & \tiny\%$ \uparrow$ & \tiny W$ 1$$ \downarrow$ & \tiny TV$ \downarrow$ & \tiny TV$ \downarrow$ & \tiny len\;W$ 1$$ \downarrow$ & \tiny ang\;W$ 1$$ \downarrow$ \\

VoxMol - Unrotated Input & 72 \quad\tiny{($\pm$29)} & 99 \quad\tiny{($\pm$1)} & 81 \quad\tiny{($\pm$31)} & 57 \quad\tiny{($\pm$20)} & 28 \quad\tiny{($\pm$3)} & 0.188 \quad\tiny{($\pm$0.071)} & 0.058 \quad\tiny{($\pm$0.030)} & 0.005 \quad\tiny{($\pm$0.001)} & 3.567 \quad\tiny{($\pm$1.348)} \\
VoxMol - Rotated Input & 71 \quad\tiny{($\pm$29)} & 99 \quad\tiny{($\pm$1)} & 81 \quad\tiny{($\pm$31)} & 57 \quad\tiny{($\pm$20)} & 28 \quad\tiny{($\pm$3)} & 0.187 \quad\tiny{($\pm$0.071)} & 0.058 \quad\tiny{($\pm$0.030)} & 0.005 \quad\tiny{($\pm$0.001)} & 3.528 \quad\tiny{($\pm$1.250)} \\
VoxMol (28 M) - Unrotated Input & 67 \quad\tiny{($\pm$22)} & 99 \quad\tiny{($\pm$1)} & 84 \quad\tiny{($\pm$21)} & 77 \quad\tiny{($\pm$17)} & 26 \quad\tiny{($\pm$2)} & 0.169 \quad\tiny{($\pm$0.064)} & 0.057 \quad\tiny{($\pm$0.030)} & 0.004 \quad\tiny{($\pm$0.001)} & 3.244 \quad\tiny{($\pm$1.199)} \\
VoxMol (28 M) - Rotated Input & 64 \quad\tiny{($\pm$21)} & 99 \quad\tiny{($\pm$1)} & 84 \quad\tiny{($\pm$21)} & 81 \quad\tiny{($\pm$15)} & 26 \quad\tiny{($\pm$2)} & 0.163 \quad\tiny{($\pm$0.064)} & 0.055 \quad\tiny{($\pm$0.029)} & 0.004 \quad\tiny{($\pm$0.001)} & 3.154 \quad\tiny{($\pm$1.204)} \\
VoxMol (7 M) - Unrotated Input & 53 \quad\tiny{($\pm$18)} & 98 \quad\tiny{($\pm$1)} & 84 \quad\tiny{($\pm$15)} & 90 \quad\tiny{($\pm$11)} & 29 \quad\tiny{($\pm$2)} & 0.165 \quad\tiny{($\pm$0.060)} & 0.059 \quad\tiny{($\pm$0.030)} & 0.004 \quad\tiny{($\pm$0.001)} & 3.469 \quad\tiny{($\pm$1.029)} \\
VoxMol (7 M) - Rotated Input & 49 \quad\tiny{($\pm$18)} & 98 \quad\tiny{($\pm$1)} & 84 \quad\tiny{($\pm$15)} & 92 \quad\tiny{($\pm$10)} & 29 \quad\tiny{($\pm$2)} & 0.159 \quad\tiny{($\pm$0.059)} & 0.057 \quad\tiny{($\pm$0.029)} & 0.004 \quad\tiny{($\pm$0.001)} & 3.424 \quad\tiny{($\pm$0.989)} \\

\end{tabular}
\caption{\textbf{Effect of model size} on seeded generation results with models initialized with lead molecules without rotations (Brownian Motion (``BM")) and with rotations (``rot."). Models were trained on 50 \% training subset.}
\label{tab:gen midi overlap2}
\vskip -0.2in
\end{table*}

\begin{table*}
\begin{tabular}{l | x{25}x{25}x{25}x{25}x{25}x{25}x{25}x{25}x{25}}
  & \tiny stable & \tiny stable & \tiny valid & \tiny unique & \tiny valency & \tiny atom & \tiny bond & \tiny bond & \tiny bond \\
  & \tiny mol. $ \uparrow$ & \tiny atom $ \uparrow$ & \tiny\%$ \uparrow$ & \tiny\%$ \uparrow$ & \tiny W$ 1$$ \downarrow$ & \tiny TV$ \downarrow$ & \tiny TV$ \downarrow$ & \tiny len\;W$ 1$$ \downarrow$ & \tiny ang\;W$ 1$$ \downarrow$ \\

VoxMol 1 \%  - Unrotated Input & 36 \quad\tiny{($\pm$16)} & 97 \quad\tiny{($\pm$1)} & 85 \quad\tiny{($\pm$13)} & 98 \quad\tiny{($\pm$5)} & 29 \quad\tiny{($\pm$2)} & 0.171 \quad\tiny{($\pm$0.053)} & 0.067 \quad\tiny{($\pm$0.037)} & 0.004 \quad\tiny{($\pm$0.001)} & 5.800 \quad\tiny{($\pm$0.890)} \\

VoxMol 1 \% - Rotated Input & 34 \quad\tiny{($\pm$15)} & 96 \quad\tiny{($\pm$1)} & 84 \quad\tiny{($\pm$14)} & 98 \quad\tiny{($\pm$5)} & 29 \quad\tiny{($\pm$2)} & 0.169 \quad\tiny{($\pm$0.052)} & 0.067 \quad\tiny{($\pm$0.036)} & 0.004 \quad\tiny{($\pm$0.001)} & 5.805 \quad\tiny{($\pm$0.896)} \\
\hline
VoxMol 10 \%  - Unrotated Input & 73 \quad\tiny{($\pm$27)} & 99 \quad\tiny{($\pm$1)} & 84 \quad\tiny{($\pm$27)} & 58 \quad\tiny{($\pm$20)} & 27 \quad\tiny{($\pm$3)} & 0.183 \quad\tiny{($\pm$0.068)} & 0.058 \quad\tiny{($\pm$0.030)} & 0.005 \quad\tiny{($\pm$0.001)} & 3.745 \quad\tiny{($\pm$1.576)} \\

VoxMol 10 \% - Rotated Input & 69 \quad\tiny{($\pm$26)} & 99 \quad\tiny{($\pm$1)} & 84 \quad\tiny{($\pm$26)} & 64 \quad\tiny{($\pm$19)} & 27 \quad\tiny{($\pm$2)} & 0.175 \quad\tiny{($\pm$0.069)} & 0.056 \quad\tiny{($\pm$0.029)} & 0.005 \quad\tiny{($\pm$0.001)} & 3.487 \quad\tiny{($\pm$1.420)} \\
\hline
VoxMol 25 \%  - Unrotated Input & 76 \quad\tiny{($\pm$30)} & 99 \quad\tiny{($\pm$1)} & 84 \quad\tiny{($\pm$29)} & 44 \quad\tiny{($\pm$19)} & 28 \quad\tiny{($\pm$3)} & 0.195 \quad\tiny{($\pm$0.075)} & 0.058 \quad\tiny{($\pm$0.031)} & 0.005 \quad\tiny{($\pm$0.001)} & 4.055 \quad\tiny{($\pm$1.786)} \\

VoxMol 25 \%  - Rotated Input & 72 \quad\tiny{($\pm$29)} & 99 \quad\tiny{($\pm$1)} & 84 \quad\tiny{($\pm$28)} & 51 \quad\tiny{($\pm$20)} & 27 \quad\tiny{($\pm$3)} & 0.185 \quad\tiny{($\pm$0.075)} & 0.057 \quad\tiny{($\pm$0.030)} & 0.005 \quad\tiny{($\pm$0.001)} & 3.649 \quad\tiny{($\pm$1.503)} \\
\hline
VoxMol 50 \% - Unrotated Input & 75 \quad\tiny{($\pm$30)} & 99 \quad\tiny{($\pm$1)} & 83 \quad\tiny{($\pm$30)} & 46 \quad\tiny{($\pm$20)} & 28 \quad\tiny{($\pm$4)} & 0.196 \quad\tiny{($\pm$0.074)} & 0.058 \quad\tiny{($\pm$0.031)} & 0.005 \quad\tiny{($\pm$0.002)} & 3.944 \quad\tiny{($\pm$1.677)} \\

VoxMol 50 \% - Rotated Input & 70 \quad\tiny{($\pm$29)} & 99 \quad\tiny{($\pm$1)} & 84 \quad\tiny{($\pm$28)} & 53 \quad\tiny{($\pm$19)} & 27 \quad\tiny{($\pm$3)} & 0.186 \quad\tiny{($\pm$0.075)} & 0.056 \quad\tiny{($\pm$0.030)} & 0.005 \quad\tiny{($\pm$0.001)} & 3.531 \quad\tiny{($\pm$1.400)} \\

\end{tabular}
\caption{\textbf{Effect of training set size} on seeded generation results with models initialized with lead molecules without rotations (Brownian Motion (``BM")) and with rotations (``rot."). Models were trained on 50 \% training subset.}
\label{tab:gen midi overlap3}
\vskip -0.2in
\end{table*}

\begin{table*}
\begin{tabular}{l | x{25}x{25}x{25}x{25}x{25}x{25}x{25}x{25}x{25}}
  & \tiny stable & \tiny stable & \tiny valid & \tiny unique & \tiny valency & \tiny atom & \tiny bond & \tiny bond & \tiny bond \\
  & \tiny mol. $ \uparrow$ & \tiny atom $ \uparrow$ & \tiny\%$ \uparrow$ & \tiny\%$ \uparrow$ & \tiny W$ 1$$ \downarrow$ & \tiny TV$ \downarrow$ & \tiny TV$ \downarrow$ & \tiny len\;W$ 1$$ \downarrow$ & \tiny ang\;W$ 1$$ \downarrow$ \\

VoxMol 50 epochs - Unrotated Input & 63 \quad\tiny{($\pm$20)} & 99 \quad\tiny{($\pm$1)} & 85 \quad\tiny{($\pm$18)} & 84 \quad\tiny{($\pm$14)} & 25 \quad\tiny{($\pm$2)} & 0.158 \quad\tiny{($\pm$0.056)} & 0.057 \quad\tiny{($\pm$0.030)} & 0.004 \quad\tiny{($\pm$0.001)} & 3.250 \quad\tiny{($\pm$1.209)} \\

VoxMol 50 epochs - Rotated Input & 60 \quad\tiny{($\pm$20)} & 99 \quad\tiny{($\pm$1)} & 85 \quad\tiny{($\pm$18)} & 87 \quad\tiny{($\pm$12)} & 25 \quad\tiny{($\pm$2)} & 0.153 \quad\tiny{($\pm$0.056)} & 0.055 \quad\tiny{($\pm$0.028)} & 0.004 \quad\tiny{($\pm$0.001)} & 3.193 \quad\tiny{($\pm$1.281)} \\
\hline
VoxMol 100 epochs - Unrotated Input & 69 \quad\tiny{($\pm$23)} & 99 \quad\tiny{($\pm$1)} & 84 \quad\tiny{($\pm$23)} & 73 \quad\tiny{($\pm$18)} & 26 \quad\tiny{($\pm$3)} & 0.169 \quad\tiny{($\pm$0.062)} & 0.057 \quad\tiny{($\pm$0.030)} & 0.004 \quad\tiny{($\pm$0.001)} & 3.330 \quad\tiny{($\pm$1.348)} \\

VoxMol 100 epochs - Rotated Input & 66 \quad\tiny{($\pm$23)} & 99 \quad\tiny{($\pm$1)} & 84 \quad\tiny{($\pm$22)} & 76 \quad\tiny{($\pm$16)} & 26 \quad\tiny{($\pm$2)} & 0.164 \quad\tiny{($\pm$0.064)} & 0.056 \quad\tiny{($\pm$0.029)} & 0.004 \quad\tiny{($\pm$0.001)} & 3.165 \quad\tiny{($\pm$1.221)} \\
\hline
VoxMol 200 epochs - Unrotated Input & 70 \quad\tiny{($\pm$28)} & 99 \quad\tiny{($\pm$1)} & 84 \quad\tiny{($\pm$28)} & 61 \quad\tiny{($\pm$20)} & 27 \quad\tiny{($\pm$3)} & 0.181 \quad\tiny{($\pm$0.065)} & 0.057 \quad\tiny{($\pm$0.029)} & 0.005 \quad\tiny{($\pm$0.001)} & 3.459 \quad\tiny{($\pm$1.376)} \\

VoxMol 200 epochs - Rotated Input & 66 \quad\tiny{($\pm$26)} & 99 \quad\tiny{($\pm$1)} & 83 \quad\tiny{($\pm$26)} & 67 \quad\tiny{($\pm$18)} & 27 \quad\tiny{($\pm$2)} & 0.173 \quad\tiny{($\pm$0.067)} & 0.055 \quad\tiny{($\pm$0.028)} & 0.004 \quad\tiny{($\pm$0.001)} & 3.290 \quad\tiny{($\pm$1.343)} \\
\hline
VoxMol 300 epochs - Unrotated Input & 73 \quad\tiny{($\pm$29)} & 99 \quad\tiny{($\pm$1)} & 83 \quad\tiny{($\pm$29)} & 54 \quad\tiny{($\pm$21)} & 27 \quad\tiny{($\pm$3)} & 0.187 \quad\tiny{($\pm$0.070)} & 0.058 \quad\tiny{($\pm$0.030)} & 0.005 \quad\tiny{($\pm$0.001)} & 3.737 \quad\tiny{($\pm$1.601)} \\

VoxMol 300 epochs - Rotated Input & 67 \quad\tiny{($\pm$28)} & 99 \quad\tiny{($\pm$1)} & 83 \quad\tiny{($\pm$28)} & 61 \quad\tiny{($\pm$20)} & 27 \quad\tiny{($\pm$2)} & 0.178 \quad\tiny{($\pm$0.072)} & 0.056 \quad\tiny{($\pm$0.029)} & 0.005 \quad\tiny{($\pm$0.001)} & 3.415 \quad\tiny{($\pm$1.394)} \\
\hline
VoxMol 400 epochs - Unrotated Input & 75 \quad\tiny{($\pm$28)} & 99 \quad\tiny{($\pm$1)} & 85 \quad\tiny{($\pm$28)} & 50 \quad\tiny{($\pm$21)} & 28 \quad\tiny{($\pm$4)} & 0.189 \quad\tiny{($\pm$0.070)} & 0.058 \quad\tiny{($\pm$0.030)} & 0.005 \quad\tiny{($\pm$0.002)} & 3.935 \quad\tiny{($\pm$1.680)} \\

VoxMol 400 epochs - Rotated Input & 70 \quad\tiny{($\pm$28)} & 99 \quad\tiny{($\pm$1)} & 85 \quad\tiny{($\pm$26)} & 57 \quad\tiny{($\pm$20)} & 27 \quad\tiny{($\pm$3)} & 0.179 \quad\tiny{($\pm$0.070)} & 0.056 \quad\tiny{($\pm$0.030)} & 0.005 \quad\tiny{($\pm$0.001)} & 3.496 \quad\tiny{($\pm$1.381)} \\
\hline
VoxMol 500 epochs - Unrotated Input & 76 \quad\tiny{($\pm$29)} & 99 \quad\tiny{($\pm$1)} & 84 \quad\tiny{($\pm$29)} & 47 \quad\tiny{($\pm$20)} & 28 \quad\tiny{($\pm$4)} & 0.194 \quad\tiny{($\pm$0.073)} & 0.058 \quad\tiny{($\pm$0.031)} & 0.005 \quad\tiny{($\pm$0.002)} & 3.968 \quad\tiny{($\pm$1.871)} \\

VoxMol 500 epochs - Rotated Input & 69 \quad\tiny{($\pm$29)} & 99 \quad\tiny{($\pm$1)} & 84 \quad\tiny{($\pm$28)} & 53 \quad\tiny{($\pm$19)} & 27 \quad\tiny{($\pm$3)} & 0.184 \quad\tiny{($\pm$0.074)} & 0.056 \quad\tiny{($\pm$0.030)} & 0.005 \quad\tiny{($\pm$0.001)} & 3.519 \quad\tiny{($\pm$1.476)} \\
\end{tabular}
\caption{\textbf{Effect of number of training epochs} on seeded generation results with models initialized with lead molecules without rotations (Brownian Motion (``BM")) and with rotations (``rot."). Models were trained on 50 \% training subset.}
\label{tab:gen midi overlap4}
\vskip -0.2in
\end{table*}

\end{document}